%% file: arxiv.tex
\documentclass{mosi}
\usepackage[utf8]{inputenc}
\usepackage{algorithm,algpseudocode}
\usepackage{url,xurl}
\setcitestyle{authoryear,round,citesep={;},aysep={,},yysep={;}}
\counterwithout{theorem}{section}

\hypersetup{
  pdftitle={ORPG: Reconciling Multiple Reward Objectives through Objective-wise Policy Gradients},
  pdfauthor={Shicheng Fang, Yiwen Zhao, Wenbo Tian, Jiahao Lu, Yining Zheng, Yuxin Wang, Xipeng Qiu}
}

\newcommand{\result}[2]{\ensuremath{\text{#1}_{\text{\normalfont\tiny\ensuremath{\pm}#2}}}}
\title{ORPG: Reconciling Multiple Reward\\Objectives through Objective-wise\\Policy Gradients}
\author{Shicheng~Fang$^{1,2,*}$, Yiwen~Zhao$^{1,*}$, Wenbo~Tian$^{1,*}$, Jiahao~Lu$^{1,2}$, Yining~Zheng$^{1,2,\dagger}$, Yuxin~Wang$^{1,2,\dagger}$, Xipeng~Qiu$^{1,2,\dagger}$}
\affil[1]{Fudan University}
\affil[2]{Shanghai Innovation Institute}
\authormark{$^{*}$Equal contribution\quad$^{\dagger}$Corresponding author}
\abstract{\input{paper/sections/abstract}}
\checkdata[Code]{\url{https://github.com/euReKa025/ORPG}}
\begin{document}
\maketitle
\input{paper/sections/introduction}
\input{paper/sections/related_work}
\input{paper/sections/method}
\input{paper/sections/experiments}
\input{paper/sections/conclusion}
\clearpage
B
\bibliography{paper/references}
\bibliographystyle{iclr2027_conference}
\clearpage
\appendix
\input{paper/sections/appendix}
\end{document}

%% file: paper/sections/abstract.tex
Multi-reward policy optimization requires a joint update that reflects both the learning signals and the intended relationships among objectives. We introduce Objective-wise Reconciled Policy Gradient (ORPG), which constructs a separate clipped policy objective for each reward and reconciles the resulting gradients into one policy update. For compatible gradients, a cosine-dependent interpolation coordinates their contributions through a partially normalized reference while preserving the norm of their sum. We characterize this update as the unique solution of a spherical directional compromise. For conflicting gradients, projection follows the task's priorities. We evaluate the same compatible rule in helpfulness--safety alignment and correctness--cost optimization for mathematical reasoning. ORPG substantially improves average Useful and Harmless scores over the strongest external baseline on each axis. In mathematics, it achieves the highest average full-budget accuracy and three-budget hypervolume among the compared methods, with more accurate and shorter responses than the initial policy. Component comparisons and training dynamics show the larger contribution of compatible coordination and a complementary benefit from conflict handling. These results support gradient reconciliation for objectives with equal standing and for objectives with an explicit priority.

%% file: paper/sections/introduction.tex
\section{Introduction}
Language models are increasingly expected to satisfy several requirements within the same response. Helpfulness and safety jointly shape the behavior of an assistant \citep{ouyang2022instructgpt,dai2023saferlhf}, while mathematical reasoning requires accurate answers at a manageable generation cost \citep{aggarwal2025l1,liu2025laser}. Reinforcement learning provides a way to train for these requirements through separate reward signals. These signals guide the same policy, so learning from them involves deciding how each objective contributes to a shared update. The desired relationship between objectives also depends on the task: helpfulness and safety may have equal standing, whereas reducing generation cost should remain subordinate to correctness. Multi-reward policy optimization therefore requires a way to coordinate learning across objectives while respecting these relationships.

Existing multi-reward optimizers combine objectives at different stages. GRPO-based methods can combine rewards before group-relative advantage estimation \citep{shao2024deepseekmath}. MO-GRPO and GDPO retain reward-specific statistics when constructing advantages \citep{ichihara2025mogrpo,liu2026gdpo}, while GD$^2$PO further processes interactions among reward-wise advantages \citep{liu2026gd2po}. Gradient-based approaches, including GAPO and PAMA, make the shared update direction an explicit design choice \citep{li-etal-2025-gradient,he2025pama}. Separate gradients create a second design problem: how should they form one update? Their directions determine whether contributions reinforce or oppose one another, while their relative norms determine their amplitudes in a direct sum. Conflict projection alone leaves compatible pairs unchanged \citep{yu2020pcgrad}, even when one gradient dominates that sum. This motivates coordinating compatible contributions alongside resolving conflicts according to task priorities.

We propose Objective-wise Reconciled Policy Gradient (ORPG), which retains a separate clipped policy objective for each reward and computes its gradient over the same trainable parameters. For a compatible pair, ORPG forms a reference by partially normalizing gradient magnitudes and interpolates between this reference and the original gradient sum. The mixing strength depends on gradient cosine similarity, and a final normalization preserves the sum's norm. For a conflicting pair, ORPG applies symmetric projection to objectives with equal standing, or a one-sided projection that preserves the primary gradient. The same compatible rule therefore serves different conflict priorities. We characterize the compatible update as the unique solution of a directional compromise on a fixed-radius sphere.

We evaluate ORPG on helpfulness--safety alignment and correctness--cost optimization for mathematical reasoning. ORPG improves both Useful and Harmless scores across all three alignment datasets, with average gains of 0.415 and 0.446 over the strongest external baseline on each axis. In mathematics, it achieves the highest average full-budget accuracy and three-budget hypervolume among the compared methods, while improving accuracy and reducing response length relative to the initial policy. Component comparisons in helpfulness--safety identify compatible coordination as the larger source of improvement, with conflict handling providing an additional benefit. Training measurements connect these gains to stronger joint reward learning and the gradient relationships encountered during optimization.

Our contributions are:
\begin{itemize}
\item We introduce ORPG, which preserves separate clipped policy objectives and reconciles their full gradients through compatible contribution coordination and task-priority conflict resolution.
\item We show that the compatible rule uniquely solves a spherical directional compromise, preserves the original sum norm, and bounds the unit-direction contribution ratio between the original and partially normalized ratios.
\item We demonstrate joint helpfulness--safety gains and accuracy-prioritized cost reduction in mathematical reasoning. Component comparisons and training measurements identify the roles of compatible coordination and conflict handling.
\end{itemize}

%% file: paper/sections/related_work.tex
\section{Related Work}
\paragraph{Multi-objective gradient coordination.}
Gradient-based multi-task methods provide several ways to construct a shared update. GradNorm adapts task weights using gradient magnitudes and relative training rates \citep{chen2018gradnorm}. MGDA-based multi-task learning seeks a common descent direction through a combination of objective gradients \citep{sener2018multi}, while gradient similarity can regulate auxiliary updates relative to a primary objective \citep{du2018auxiliary}. PCGrad projects conflicting gradients \citep{yu2020pcgrad}. GradVac adjusts gradient relationships toward target cosine similarities \citep{wang2021gradvac}, and CAGrad controls the worst local objective improvement around the average gradient \citep{liu2021cagrad}. Aligned-MTL constructs the update through an alignment-based transformation of the gradient system \citep{senushkin2023aligned}. These approaches differ in the quantity they control: task weights, local directional improvement, pairwise relationships, or the conditioning of the gradient system.

\paragraph{Reinforcement learning and multi-reward policy optimization.}
PPO introduced a clipped policy objective for stable policy updates \citep{schulman2017proximalpolicyoptimizationalgorithms}. GRPO estimates advantages from groups of sampled responses and removes the need for a learned value function \citep{shao2024deepseekmath}. Subsequent multi-objective alignment methods extended the construction of the update. GAPO rescales objective gradients and solves a minimum-norm combination problem \citep{li-etal-2025-gradient}, while PAMA combines a modified policy objective with efficient multi-objective weight calculation \citep{he2025pama}. Dynamic reward weighting adapts objective weights during training \citep{lu2025dynamicreward}. MO-GRPO and GDPO normalize rewards separately before aggregating their advantages \citep{ichihara2025mogrpo,liu2026gdpo}. Blockwise advantage estimation assigns objective-specific signals to corresponding response blocks \citep{pavlenko2026blockwise}, and GD$^2$PO filters conflicting reward-wise advantages and reweights prompt groups \citep{liu2026gd2po}. The combination stage determines which interactions the optimizer can act on explicitly. Reward and advantage methods shape the learning signal before policy differentiation; gradient methods operate on the parameter update induced by that signal.\citep{Li2026RethinkingTR,Zhao2026TowardsBA} ORPG preserves each reward through a separate clipped policy objective and reconciles the resulting full policy gradients into a joint update.

Length-aware reasoning methods express generation cost through length targets, penalties, or response selection \citep{aggarwal2025l1,luo2025o1pruner,yi2025shorterbetter,liu2025dler,shrivastava2025gfpo,liu2025laser}. This setting gives the objectives a primary--secondary relationship: correctness determines answer quality, while length controls the cost of obtaining it.

%% file: paper/sections/method.tex
\section{Objective-wise Reconciled Policy Gradient}
\label{sec:method}
\begin{figure}[t]
\centering
\includegraphics[width=\linewidth]{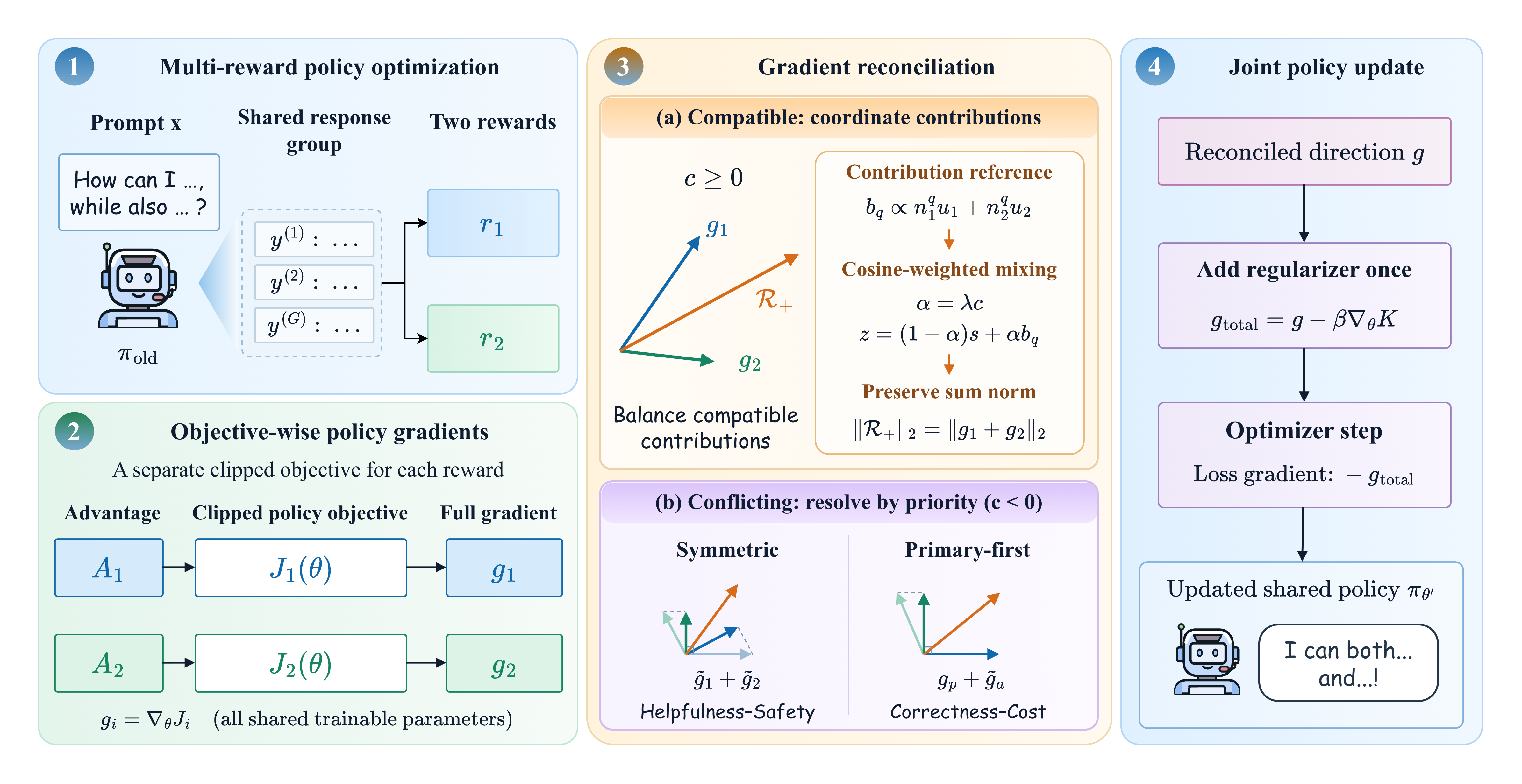}
\caption{Overview of ORPG. Each reward retains its own clipped policy objective. Reconciliation uses the relationship between their full policy gradients to form a joint update.}
\label{fig:method}
\end{figure}
\subsection{Policy optimization setup and separate objectives}
Let $\pi_\theta$ be a policy with trainable parameters $\theta$. For each prompt $x\sim\mathcal D$, a fixed rollout policy $\pi_{\mathrm{old}}$ samples a group of $G$ responses $y^{(1)},\ldots,y^{(G)}$. Reward $i$ assigns each response an advantage $A_i^{(j)}$, broadcast over its valid response tokens. GRPO constructs these advantages from within-group reward statistics \citep{shao2024deepseekmath}; the task-specific constructions used here are given in Section~\ref{sec:experiments} and Appendix~\ref{app:implementation}.

For token $t$ of response $j$, define the importance ratio
\begin{equation}
\rho_{j,t}(\theta)=\frac{\pi_\theta(y_t^{(j)}\mid x,y_{<t}^{(j)})}{\pi_{\mathrm{old}}(y_t^{(j)}\mid x,y_{<t}^{(j)})}.
\end{equation}
PPO-style clipping, also used by GRPO, gives the maximized surrogate integrand \citep{schulman2017proximalpolicyoptimizationalgorithms,shao2024deepseekmath}
\begin{equation}
\phi_{\mathrm{clip}}(\rho,A)=\min\{\rho A,\operatorname{clip}(\rho,1-\epsilon_-,1+\epsilon_+)A\}.
\end{equation}
ORPG retains a separate objective for each reward:
\begin{equation}
J_i(\theta)=\mathbb E_{x\sim\mathcal D,\,\{y^{(j)}\}\sim\pi_{\mathrm{old}}}\!\left[\operatorname{Reduce}_{j,t}\phi_{\mathrm{base}}(\rho_{j,t}(\theta),A_i^{(j)})\right].
\label{eq:objectives}
\end{equation}
Here $\phi_{\mathrm{base}}$ includes the negative-advantage safeguard specified in Appendix~\ref{app:implementation}. The reduction is a valid-token mean for helpfulness--safety and a sequence mean of token means for mathematics. Separate clipping preserves objective identity through differentiation.

We use ascent notation $g_i=\nabla_\theta J_i$, with each gradient covering all trainable policy parameters. The reconciliation operator $\mathcal R$ combines these gradients, followed by one shared regularizer:
\begin{equation}
g_{\mathrm{total}}=\mathcal R(g_1,\ldots,g_m)-\beta\nabla_\theta K(\theta).
\label{eq:regularizer}
\end{equation}
The optimizer uses $-g_{\mathrm{total}}$ as its loss gradient. The framework supports multiple rewards; our implemented and evaluated rule treats two.

\subsection{Compatible contributions}
For two nonzero gradients, write
\begin{equation}
n_i=\|g_i\|_2,\qquad u_i=g_i/n_i,\qquad c=u_1^\top u_2,\qquad s=g_1+g_2,\qquad S=\|s\|_2.
\end{equation}
When $c\geq0$, both gradients are locally compatible. Their relative norms still determine their amplitudes in $s=n_1u_1+n_2u_2$. ORPG coordinates these amplitudes using a partially normalized reference:
\begin{equation}
v_q=n_1^q u_1+n_2^q u_2,\qquad b_q=S\frac{v_q}{\|v_q\|_2},\qquad q\in[0,1].
\label{eq:reference}
\end{equation}
The choice $q=1$ recovers the original sum direction, while $q=0$ gives equal amplitudes on the unit directions. At $q=\tfrac12$, their ratio becomes $\sqrt{n_1/n_2}$, halfway between equal amplitudes and the original ratio in logarithmic coordinates. This retains information about gradient magnitude while moderating its influence on the joint direction.

Let $\lambda\in[0,1]$ control the maximum mixing strength and set $\alpha=\lambda c$. The compatible update is
\begin{equation}
z=(1-\alpha)s+\alpha b_q,\qquad
\mathcal{R}_+(g_1,g_2)=S\frac{z}{\|z\|_2}.
\label{eq:compatible}
\end{equation}
Cosine similarity controls how strongly the reference contributes. Near orthogonality, the adjustment approaches zero. The final normalization retains the magnitude of the original gradient sum while changing its direction.

\begin{proposition}[Spherical directional compromise]
\label{prop:sphere}
For nonzero $g_1,g_2$ with $c\geq0$, $q\in[0,1]$, and $\lambda\in[0,1]$, Equation~\eqref{eq:compatible} is the unique solution of
\begin{equation}
\underset{\|g\|_2=S}{\operatorname{minimize}}\quad
(1-\alpha)\|g-s\|_2^2+\alpha\|g-b_q\|_2^2.
\label{eq:sphere}
\end{equation}
\end{proposition}
The objective balances proximity to the original sum and to the contribution reference on the same sphere. Expanding the squares reduces the problem to maximizing $g^\top z$ under a norm constraint. Its solution is the normalized vector in Equation~\eqref{eq:compatible}; a full derivation appears in Appendix~\ref{app:proof}.

Several properties follow directly. The output equals $s$ when $\lambda=0$, $c=0$, $q=1$, the gradients have equal norms, or they point in the same direction. For an unequal-norm pair, the coefficient ratio after mixing lies between the original ratio and its $q$-power reference. The adjustment therefore changes contributions continuously rather than replacing gradient magnitudes with a binary choice.

\subsection{Conflict resolution and priorities}
When $d=g_1^\top g_2<0$, ORPG uses a conflict rule determined by the task priorities. With symmetric objectives, it applies the two-objective PCGrad projection \citep{yu2020pcgrad}:
\begin{equation}
\widetilde g_1=g_1-\frac{d}{n_2^2}g_2,\qquad
\widetilde g_2=g_2-\frac{d}{n_1^2}g_1,\qquad
\mathcal{R}_-=\widetilde g_1+\widetilde g_2.
\label{eq:symmetric}
\end{equation}
Each projected direction removes its component opposing the other objective.

For a primary objective $p$ and a secondary objective $a$, ORPG preserves $g_p$ and finds the closest secondary direction that does not oppose it:
\begin{equation}
\widetilde g_a=\arg\min_h\frac12\|h-g_a\|_2^2
\quad\text{subject to}\quad g_p^\top h\geq0.
\end{equation}
For a conflicting pair, the closed-form result is
\begin{equation}
\widetilde g_a=g_a-\frac{g_p^\top g_a}{\|g_p\|_2^2}g_p,
\qquad \mathcal{R}_-=g_p+\widetilde g_a.
\label{eq:priority}
\end{equation}
The secondary objective retains its orthogonal component, while the primary direction remains intact. In ascent notation, $g_p^\top\mathcal{R}_-=\|g_p\|_2^2$ for the conflicting pair. This first-order property concerns the reconciled policy direction before shared regularization and the optimizer update.

\subsection{Overall update and optimization procedure}
At optimization step $k$, let $\mathcal B_k$ be the current minibatch and $g_i^k=\nabla_\theta J_i(\theta;\mathcal B_k)|_{\theta=\theta_k}$. The reconciliation rules determine scalar coefficients $\omega_i^k$ such that
\begin{equation}
g_{\mathrm{rec}}^k=\mathcal R(g_1^k,g_2^k)=\sum_{i=1}^2\omega_i^k g_i^k.
\label{eq:weighted-gradient}
\end{equation}
Holding these coefficients fixed for the current differentiation gives the local surrogate
\begin{equation}
\widetilde J_{\mathrm{ORPG}}^k(\theta)=\sum_{i=1}^2\operatorname{sg}(\omega_i^k)J_i(\theta;\mathcal B_k)-\beta K(\theta;\mathcal B_k),
\label{eq:local-surrogate}
\end{equation}
where $\operatorname{sg}$ denotes stop-gradient. Consequently,
\begin{equation}
\left.\nabla_\theta\widetilde J_{\mathrm{ORPG}}^k(\theta)\right|_{\theta=\theta_k}=g_{\mathrm{rec}}^k-\beta\nabla_\theta K(\theta_k;\mathcal B_k)=g_{\mathrm{total}}^k.
\label{eq:surrogate-gradient}
\end{equation}
\noindent\begin{minipage}[t]{0.47\linewidth}
\vspace{0pt}
This representation connects the reconciled direction to the reward-specific objectives. The coefficients are recomputed from the current gradients at every optimization minibatch and remain fixed only for that differentiation. Appendix~\ref{app:algorithm} gives their closed forms.

\medskip
Algorithm~\ref{alg:orpg} obtains each full objective gradient separately. Reconciliation then uses three global Gram scalars, $n_1^2,n_2^2,d$, and $O(P)$ vector operations for $P$ trainable parameters. Appendix~\ref{app:cost} details operations and training costs.

\medskip
Both symmetric projections use the original gradient pair. When either gradient is zero, the remaining objective passes through unchanged. The shared regularizer is differentiated separately and included once, with $g_K=0$ when disabled. The optimizer then clips the total loss gradient and applies AdamW. Thus, policy-objective clipping, gradient reconciliation, and final gradient-norm clipping act at distinct stages.

\medskip
The rollout policy stays fixed within each rollout batch. Both task settings use the same compatible rule; their advantage construction and conflict priority determine how the objectives enter the update.
\end{minipage}\hfill
\begin{minipage}[t]{0.50\linewidth}
\vspace{0pt}
\begin{algorithm}[H]
\caption{ORPG update}\label{alg:orpg}
\fontsize{9}{10.5}\selectfont
\begin{algorithmic}[1]
\Require $\pi_\theta$, $r_1,r_2$, $q,\lambda,\beta$, task rules
\For{each rollout batch}
\State Fix $\pi_{\mathrm{old}}$; sample response groups
\State Evaluate rewards; construct $A_1,A_2$
\For{each optimization minibatch}
\State Form separate clipped $J_1,J_2$
\State $g_i\gets\nabla J_i$; $g_K\gets\nabla K$
\State Compute global $n_1^2,n_2^2,d$
\If{$n_1n_2=0$}
\State $g\gets g_1+g_2$
\ElsIf{$d\geq0$}
\State $c\gets d/(n_1n_2)$; $\alpha\gets\lambda c$
\State $g\gets\mathcal R_+(g_1,g_2)$ via Eq.~\eqref{eq:compatible}
\ElsIf{symmetric priority}
\State $\widetilde g_1\gets g_1-dg_2/n_2^2$
\State $\widetilde g_2\gets g_2-dg_1/n_1^2$
\State $g\gets\widetilde g_1+\widetilde g_2$ \Comment{Eq.~\eqref{eq:symmetric}}
\Else
\State Identify primary $g_p$, secondary $g_a$
\State $\widetilde g_a\gets g_a-dg_p/\|g_p\|_2^2$
\State $g\gets g_p+\widetilde g_a$ \Comment{Eq.~\eqref{eq:priority}}
\EndIf
\State $h\gets -(g-\beta g_K)$ \Comment{Loss gradient}
\State Clip $h$ to the maximum gradient norm
\State Update $\theta$ with AdamW using $h$
\EndFor
\EndFor
\end{algorithmic}
\end{algorithm}
\end{minipage}

%% file: paper/sections/experiments.tex
\section{Experiments}
\label{sec:experiments}
\subsection{Experimental setup}
\paragraph{Training and repeated runs.}
Both settings start from Qwen3-4B-Instruct-2507 \citep{yang2025qwen3technicalreport} and use the \texttt{verl} framework \citep{sheng2024hybridflow}. Evaluation uses 100-step policies. The default Math run accumulates 21.91 training-step hours on eight H200 GPUs, detailed in Appendix~\ref{app:cost}. Unless stated otherwise, means and sample standard deviations are computed across three independent training runs and three evaluation runs for base. Appendix~\ref{app:implementation} provides optimization settings; Appendix~\ref{app:prompts} specifies repeated-run aggregation.

\paragraph{Data, objectives, and benchmarks.}
For helpfulness--safety, we follow Safe RLHF's separation of alignment criteria \citep{dai2023saferlhf}, using the Artessay Qwen2.5-7B-SafeRLHF reward and cost models to score
helpfulness and harmlessness\citep{qwen2025qwen25technicalreport,artessayrm2026,artessaycm2026}. Training uses Alpaca with disjoint calibration and evaluation subsets \citep{alpaca}; evaluation covers Alpaca, HH-RLHF, and PKU-SafeRLHF \citep{bai2022traininghelpfulharmlessassistant,ji-etal-2025-pku}. Group-centered advantages share a scale, and conflict resolution is symmetric. For correctness--cost, training uses DeepScaleR preview prompts \citep{deepscaler2025}; evaluation covers AIME-24, AMC-22-23, MATH, Minerva-Math, and OlympiadBench \citep{h4aime2024,aimoamc2024,hendrycks2021measuringmathematicalproblemsolving,lewkowycz2022solvingquantitativereasoningproblems,he-etal-2024-olympiadbench}. The rewards are binary correctness and an indicator of length at most $\tau=4000$. ORPG and all external training baselines use these same reward definitions and length threshold. Correctness uses group-relative advantages and receives conflict priority; length advantages are centered within the correct subset and zero elsewhere. Appendices~\ref{app:implementation} and~\ref{app:prompts} detail objective construction, dataset sizes, and prompts.

\paragraph{Baselines.}
We compare ORPG with the initial model, GRPO, GDPO, and the hard variant of GD$^2$PO \citep{shao2024deepseekmath,liu2026gdpo,liu2026gd2po}. The initial policy anchors changes in task quality and generation cost. GRPO combines rewards before constructing its group-relative update. GDPO separately normalizes reward-wise advantages before aggregation, while GD$^2$PO filters conflicting reward-wise advantages and reweights prompt groups. These comparisons distinguish coordination at the learning-signal level from reconciliation of separate policy gradients. Section~\ref{sec:components} evaluates the reconciliation components and alternative gradient combination rules under the objective-wise formulation.

\paragraph{Evaluation metrics.}
For helpfulness--safety, each evaluation run generates one response for every prompt in each complete set. We report mean Useful and Harmless scores. For mathematics, accuracy estimates pass@1 from four responses per problem. Table~\ref{tab:math} reports our primary comparison: accuracy and mean length at the 8192-token budget. Table~\ref{tab:math-hv} reports hypervolume (HV), which summarizes the accuracy--cost trade-off using 2048-, 4096-, and 8192-token measurements. Shorter-budget responses are exact prefixes of the same generations. For each dataset, HV is the union area of rectangles from $(0,0)$ to accuracy--efficiency points $(a,1-\ell/8192)$, with efficiency clipped to $[0,1]$. Avg weights datasets equally after computing their metrics. Appendix~\ref{app:metrics} gives the scoring protocol, formula, and budget-specific values.

\subsection{Results}
\input{paper/tables/hs}
Table~\ref{tab:hs} shows that ORPG achieves the highest Useful and Harmless scores on all three evaluation sets. Its average Useful score of 5.589 exceeds GDPO by 0.415, while its average Harmless score of 6.904 exceeds GD$^2$PO by 0.446. Both scores improve within each dataset, covering general instructions and the two safety-oriented evaluation sets.

\input{paper/tables/math}
\input{paper/tables/math_hv}
Table~\ref{tab:math} shows the accuracy-priority outcome in mathematical reasoning. ORPG achieves 66.7\% average accuracy at the 8192-token budget, improving on Base by 1.14 percentage points while using 611 fewer tokens per response (19.3\%). It exceeds all three external training baselines in accuracy on every dataset. Relative to Base, four datasets improve and AMC-22-23 retains the same mean accuracy. The external baselines produce shorter responses than ORPG but reduce accuracy relative to Base: their average accuracies range from 59.35\% to 60.69\%. ORPG obtains its cost reduction while improving the primary correctness objective.

Table~\ref{tab:math-hv} evaluates the joint accuracy--cost outcome across the three budgets. ORPG reaches an average HV of 0.523, compared with 0.493 for Base and 0.513 for GD$^2$PO, the strongest external baseline on this metric. ORPG's leading full-budget accuracy and aggregate HV show improved correctness and joint accuracy--cost performance. Appendix~\ref{app:metrics} provides the budget-specific measurements used to compute HV.

\subsection{Component contributions}
\label{sec:components}
\begin{table}[t]
\caption{Reconciliation components and alternative gradient rules on helpfulness--safety. All rows retain separate policy objectives. Scores are averaged over the same three datasets as Table~\ref{tab:hs}.}
\label{tab:ablation}
\begin{center}
\small
\begin{tabular}{lcc}
\toprule
Update & Useful $\uparrow$ & Harmless $\uparrow$ \\
\midrule
ORPG & \result{\textbf{5.589}}{0.0004} & \result{\textbf{6.904}}{0.001} \\
Without compatible coordination & \result{5.212}{0.062} & \result{6.710}{0.032} \\
Without conflict resolution & \result{5.563}{0.0006} & \result{6.859}{0.001} \\
Without either component & \result{5.193}{0.073} & \result{6.695}{0.085} \\
\midrule
CAGrad & \result{5.183}{0.002} & \result{6.626}{0.002} \\
Aligned-MTL & \result{5.065}{0.002} & \result{6.501}{0.005} \\
\bottomrule
\end{tabular}
\end{center}
\end{table}
All variants in Table~\ref{tab:ablation} retain the same objective-wise policy structure. Without either component uses the direct gradient sum (Sum). Without compatible coordination applies conflict projection and directly sums compatible gradients (PCGrad). Without conflict resolution retains compatible coordination and directly sums opposing gradients. CAGrad and Aligned-MTL replace the reconciliation operator with their respective joint-gradient rules. Appendix~\ref{app:variants} gives the formulas.

Table~\ref{tab:ablation} shows that ORPG achieves the highest Useful and Harmless scores among the objective-wise rules. Its improvement over the version without either component establishes the benefit of coordinating the gradients after preserving separate policy objectives. Compatible coordination provides the larger component gain: removing it reduces Useful by 0.378 and Harmless by 0.193. Removing compatible coordination gives results close to removing both components, while the compatible-only variant approaches the full method. The distinction between these updates is how they combine compatible gradients, connecting the largest gain to the central design choice in ORPG.

Conflict resolution further improves both scores. ORPG also exceeds CAGrad and Aligned-MTL on both axes. Section~\ref{sec:training-dynamics} examines the learning dynamics. In mathematics, the full rule achieves the highest 8192-budget accuracy among the four component versions, exceeding the version without conflict resolution by 1.20 percentage points. Appendix~\ref{app:math-components} reports the accuracy--cost comparison.

All five configurations per setting outperform the external training baselines on both HS scores, full-budget mathematical accuracy, and HV. The default leads in HS scores and full-budget accuracy; $q=0.75$ achieves higher mathematical HV. Appendix~\ref{app:sensitivity} reports the individual $q$ and $\lambda$ scans.

\subsection{Training dynamics}
\label{sec:training-dynamics}
\begin{figure}[!ht]
\centering
\includegraphics[width=0.74\linewidth]{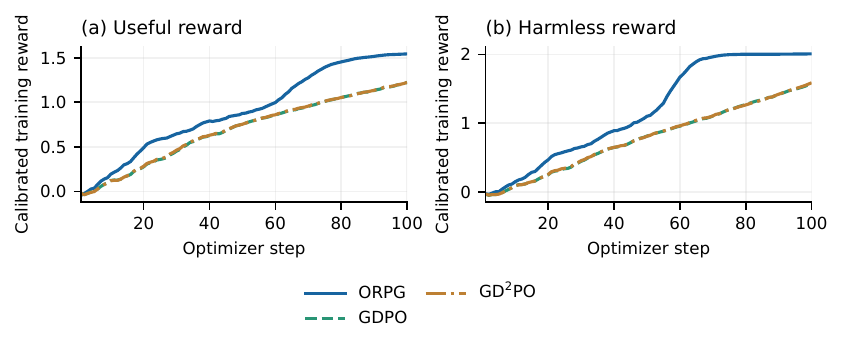}
\caption{Training rewards against external baselines. (a) Useful and (b) Harmless. A trailing five-step mean is applied to each curve. These calibrated training rewards differ from the raw evaluation scores in Table~\ref{tab:hs}.}
\label{fig:reward-external}
\end{figure}

\begin{figure}[!ht]
\centering
\includegraphics[width=0.86\linewidth]{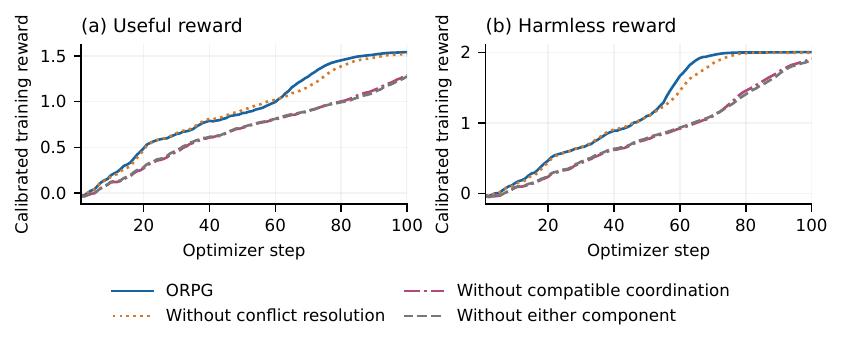}
\caption{Training rewards under objective-wise gradient rules. (a) Useful and (b) Harmless. Colors and line styles identify the same versions throughout. Smoothing matches Figure~\ref{fig:reward-external}.}
\label{fig:reward-components}
\end{figure}

ORPG learns stronger usefulness and harmlessness together during training. We compare calibrated training rewards against external policy optimizers and objective-wise gradient rules in Figures~\ref{fig:reward-external} and~\ref{fig:reward-components}. Appendix~\ref{app:reward-dynamics} gives the run-level statistics and advantage measurements.

\paragraph{Stronger joint learning than external baselines.}
Figures~\ref{fig:reward-external}(a--b) show that ORPG develops an advantage on both rewards and extends it through the middle and later stages. The separation is especially visible around steps 60--80: Useful continues to rise while Harmless reaches a higher level. Over the final 20 steps, ORPG averages 1.522 Useful and 2.005 Harmless, compared with 1.157 and 1.458 for GDPO, and 1.158 and 1.462 for GD$^2$PO. The simultaneous gains connect the stronger held-out scores to improved learning of both training objectives.

\paragraph{Compatible coordination provides the main reward gain.}
Figures~\ref{fig:reward-components}(a--b) separate the four component versions. The two retaining compatible coordination develop substantially higher rewards in the middle and later stages, while removing this component gives a trajectory close to removing both. Over the final 20 steps, ORPG reaches 1.522 Useful and 2.005 Harmless, compared with 1.492 and 1.999 without conflict resolution. Both exceed the versions without compatible coordination and without either component. Together with Table~\ref{tab:ablation}, these trajectories identify the main gain from compatible coordination and the additional improvement from the complete rule.

\begin{figure}[!ht]
\centering
\includegraphics[width=0.86\linewidth]{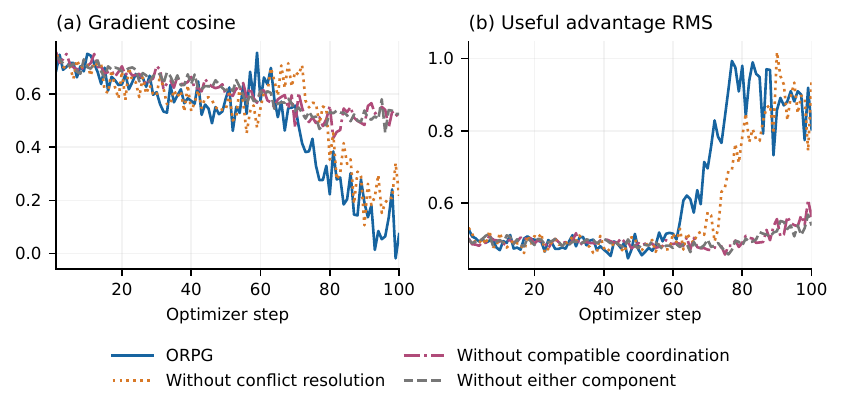}
\caption{Training signals for the four component versions. (a) Gradient cosine and (b) Useful advantage RMS, shown as unsmoothed step aggregates. Colors and line styles match Figure~\ref{fig:reward-components}.}
\label{fig:training-signals}
\end{figure}

\paragraph{Learning improves in a predominantly compatible regime.}
Figure~\ref{fig:training-signals}(a) shows positive step-average gradient cosine through most of training for all four versions. The reward gains from compatible coordination therefore develop largely in a regime where conflict projection leaves the gradient sum unchanged. This connects the training behavior to the motivation for coordinating contributions even when gradients are locally compatible. The versions without compatible coordination and without either component record zero conflicts, whereas ORPG and the version without conflict resolution encounter conflicts late in training; ORPG's average projection rate is 6.5\%. Appendix~\ref{app:telemetry} gives conflict and projection trajectories and stage summaries.

\paragraph{A stronger Useful signal accompanies reward improvement.}
Figure~\ref{fig:training-signals}(b) shows that the versions retaining compatible coordination sustain a stronger Useful advantage signal later in training. Over the final 20 steps, Useful RMS reaches 0.887 for ORPG and 0.860 without conflict resolution, compared with 0.528 without compatible coordination and 0.521 without either component. All four use the same group-centering and shared-scale advantage construction. The stronger signal accompanies the higher Useful reward in Figure~\ref{fig:reward-components}(a), indicating effective optimization of usefulness alongside the higher harmlessness reward.

%% file: paper/tables/hs.tex
\begin{table}[t]
\caption{Helpfulness--safety results. Each dataset reports Useful (U) and Harmless (H); Avg is the equal-weight average across datasets. Both scores are higher-is-better.}
\label{tab:hs}
\begin{center}
\fontsize{8}{10}\selectfont
\setlength{\tabcolsep}{2pt}
\begin{tabular}{l*{8}{c}}
\toprule
\multirow{2}{*}{Method} & \multicolumn{2}{c}{Alpaca} & \multicolumn{2}{c}{HH-RLHF} & \multicolumn{2}{c}{PKU-SafeRLHF} & \multicolumn{2}{c}{Avg} \\
\cmidrule(lr){2-3}\cmidrule(lr){4-5}\cmidrule(lr){6-7}\cmidrule(lr){8-9}
 & U & H & U & H & U & H & U & H \\
\midrule
Base & \result{2.536}{0.025} & \result{2.941}{0.028} & \result{2.855}{0.002} & \result{3.977}{0.002} & \result{4.644}{0.002} & \result{6.404}{0.004} & \result{3.345}{0.009} & \result{4.441}{0.010} \\
GRPO & \result{5.232}{0.036} & \result{6.212}{0.080} & \result{4.470}{0.037} & \result{6.082}{0.044} & \result{5.620}{0.011} & \result{6.938}{0.002} & \result{5.107}{0.028} & \result{6.411}{0.038} \\
GDPO & \result{5.335}{0.010} & \result{6.250}{0.121} & \result{4.546}{0.004} & \result{6.116}{0.019} & \result{5.640}{0.008} & \result{6.939}{0.005} & \result{5.174}{0.005} & \result{6.435}{0.047} \\
GD$^2$PO & \result{5.332}{0.010} & \result{6.297}{0.033} & \result{4.541}{0.007} & \result{6.138}{0.010} & \result{5.631}{0.001} & \result{6.939}{0.002} & \result{5.168}{0.005} & \result{6.458}{0.013} \\
\midrule
ORPG & \result{\textbf{5.943}}{0.002} & \result{\textbf{7.061}}{0.002} & \result{\textbf{5.044}}{0.002} & \result{\textbf{6.679}}{0.003} & \result{\textbf{5.781}}{0.0001} & \result{\textbf{6.972}}{0.0002} & \result{\textbf{5.589}}{0.0004} & \result{\textbf{6.904}}{0.001} \\
\bottomrule
\end{tabular}
\end{center}
\end{table}

%% file: paper/tables/math.tex
\begin{table}[t]
\caption{Mathematical accuracy (Acc, \%) and mean response length (Len, tokens) at the 8192-token budget. Each dataset is followed by the equal-weight Avg. A24: AIME-24; AMC: AMC-22-23; Min.: Minerva-Math; Oly.: OlympiadBench.}
\label{tab:math}
\begin{center}
\fontsize{7.5}{9.5}\selectfont
\setlength{\tabcolsep}{0.8pt}
\resizebox{\linewidth}{!}{%
\begin{tabular}{l*{12}{c}}
\toprule
\multirow{2}{*}{Method} & \multicolumn{2}{c}{A24} & \multicolumn{2}{c}{AMC} & \multicolumn{2}{c}{MATH} & \multicolumn{2}{c}{Min.} & \multicolumn{2}{c}{Oly.} & \multicolumn{2}{c}{Avg} \\
\cmidrule(lr){2-3}\cmidrule(lr){4-5}\cmidrule(lr){6-7}\cmidrule(lr){8-9}\cmidrule(lr){10-11}\cmidrule(lr){12-13}
 & Acc (\%) & Len & Acc (\%) & Len & Acc (\%) & Len & Acc (\%) & Len & Acc (\%) & Len & Acc (\%) & Len \\
\midrule
Base & \result{53.9}{3.4} & \result{5698}{60} & \result{\textbf{82.4}}{0.7} & \result{3351}{86} & \result{87.6}{0.1} & \result{1462}{6} & \result{37.3}{0.3} & \result{1495}{22} & \result{66.4}{0.2} & \result{3821}{9} & \result{65.5}{0.7} & \result{3165}{12} \\
GRPO & \result{42.2}{1.0} & \result{2541}{135} & \result{75.5}{2.3} & \result{1708}{58} & \result{86.0}{0.2} & \result{939}{40} & \result{37.9}{0.6} & \result{964}{91} & \result{61.8}{0.3} & \result{1688}{71} & \result{60.7}{0.5} & \result{1568}{66} \\
GDPO & \result{38.6}{1.3} & \result{\textbf{2255}}{105} & \result{74.7}{0.6} & \result{\textbf{1540}}{26} & \result{85.5}{0.1} & \result{\textbf{833}}{12} & \result{37.3}{0.6} & \result{833}{25} & \result{60.6}{0.6} & \result{\textbf{1501}}{46} & \result{59.3}{0.1} & \result{\textbf{1392}}{38} \\
GD$^2$PO & \result{41.9}{5.4} & \result{2470}{90} & \result{75.1}{1.5} & \result{1583}{51} & \result{85.4}{0.1} & \result{846}{12} & \result{37.9}{1.0} & \result{\textbf{826}}{34} & \result{60.9}{0.3} & \result{1553}{29} & \result{60.3}{0.8} & \result{1456}{29} \\
\midrule
ORPG & \result{\textbf{56.4}}{1.9} & \result{4613}{94} & \result{\textbf{82.4}}{0.6} & \result{2748}{39} & \result{\textbf{88.0}}{0.1} & \result{1229}{3} & \result{\textbf{39.3}}{0.5} & \result{1287}{27} & \result{\textbf{67.2}}{0.6} & \result{2893}{23} & \result{\textbf{66.7}}{0.3} & \result{2554}{22} \\
\bottomrule
\end{tabular}
}
\end{center}
\end{table}

%% file: paper/tables/math_hv.tex
\begin{table}[t]
\caption{Mathematical hypervolume (HV, higher is better) over the 2048-, 4096-, and 8192-token budgets. Avg gives equal weight to each dataset.}
\label{tab:math-hv}
\begin{center}
\fontsize{8}{10}\selectfont
\setlength{\tabcolsep}{3pt}
\begin{tabular}{l*{6}{c}}
\toprule
Method & A24 & AMC & MATH & Min. & Oly. & Avg \\
\midrule
Base & \result{0.280}{0.015} & \result{0.608}{0.006} & \result{0.768}{0.001} & \result{0.325}{0.002} & \result{0.485}{0.0005} & \result{0.493}{0.004} \\
GRPO & \result{0.311}{0.011} & \result{0.620}{0.014} & \result{0.769}{0.003} & \result{0.336}{0.003} & \result{0.507}{0.007} & \result{0.509}{0.003} \\
GDPO & \result{0.295}{0.011} & \result{0.625}{0.005} & \result{0.774}{0.001} & \result{0.337}{0.006} & \result{0.507}{0.006} & \result{0.507}{0.002} \\
GD$^2$PO & \result{0.315}{0.040} & \result{0.627}{0.014} & \result{0.773}{0.001} & \result{0.342}{0.009} & \result{0.508}{0.002} & \result{0.513}{0.005} \\
\midrule
ORPG & \result{\textbf{0.339}}{0.010} & \result{\textbf{0.637}}{0.004} & \result{\textbf{0.777}}{0.0005} & \result{\textbf{0.344}}{0.005} & \result{\textbf{0.516}}{0.004} & \result{\textbf{0.523}}{0.002} \\
\bottomrule
\end{tabular}
\end{center}
\end{table}

%% file: paper/sections/conclusion.tex
\section{Conclusion}
We propose ORPG, a gradient-reconciliation method for multi-reward policy optimization that improves joint performance across reward objectives. Its compatible branch coordinates relative contributions while preserving the norm of the gradient sum, and its conflict branch follows task priorities. ORPG improves usefulness and harmlessness jointly and achieves the highest average full-budget accuracy and three-budget hypervolume in mathematical reasoning among the compared methods. Component comparisons and training dynamics in helpfulness--safety identify compatible coordination as the larger source of improvement, with conflict handling adding a complementary benefit. Across the two settings, the results support coordinating objectives according to both their local gradient relationships and their task-level priorities. In summary, ORPG opens a research direction for multi-objective policy optimization through objective-wise gradient reconciliation.

%% file: paper/sections/appendix.tex
\section{Properties of the reconciliation operator}
\label{app:proof}
\subsection{Proof of Proposition~\ref{prop:sphere}}
For compatible nonzero gradients, $S>0$ and $v_q\neq0$. Moreover,
\[
s^\top v_q=n_1^{q+1}+n_2^{q+1}+c(n_1n_2^q+n_2n_1^q)>0.
\]
Thus $s^\top b_q>0$, so $z=(1-\alpha)s+\alpha b_q$ is nonzero for all $\alpha\in[0,1]$. Since $\|s\|=\|b_q\|=\|g\|=S$, the objective in Equation~\eqref{eq:sphere} is
\[
2S^2-2g^\top\big((1-\alpha)s+\alpha b_q\big)=2S^2-2g^\top z.
\]
Cauchy--Schwarz gives $g^\top z\leq S\|z\|$, with equality only at $g=Sz/\|z\|$. This proves the unique minimizer.

\subsection{Contribution ratios and identity cases}
Before the final normalization, $z=a_1u_1+a_2u_2$, where
\[
a_i=(1-\alpha)n_i+\frac{\alpha S}{\|v_q\|}n_i^q.
\]
Assume $r=n_1/n_2\geq1$. Let $w=(1-\alpha)n_2$ and $t=\alpha S n_2^q/\|v_q\|$. Then
\[
\frac{a_1}{a_2}=\frac{wr+tr^q}{w+t}\in[r^q,r].
\]
The common final normalization does not change this ratio. At $q=\tfrac12$, the reference log-ratio is half the original log-ratio. If $\alpha=0$ or $q=1$, Equation~\eqref{eq:compatible} returns $s$. Equal norms make $v_q$ proportional to $s$, as does $c=1$, so these cases also return $s$. For $q>0$, the contribution of a vanishing objective tends to zero; the implementation passes through the other gradient when one objective is inactive.

These properties describe the reconciled policy gradient. The shared regularizer and the optimizer act after reconciliation, as specified in Equation~\eqref{eq:regularizer}.

\subsection{Priority projection}
The feasible set $\{h:g_p^\top h\geq0\}$ is a closed half-space. If $g_p^\top g_a<0$, its Euclidean projection is $\widetilde g_a=g_a-(g_p^\top g_a)g_p/\|g_p\|^2$. It follows that $g_p^\top\widetilde g_a=0$, so the joint direction has primary directional derivative $\|g_p\|^2$. For a compatible pair, the operator instead uses Equation~\eqref{eq:compatible}.

\subsection{Compatible coordination for multiple objectives}
\label{app:multiobjective}
The objective-wise construction in Equation~\eqref{eq:objectives} permits any number of reward objectives. The compatible reference also admits a direct extension. Let $I$ index the nonzero gradients and suppose $g_i^\top g_j\geq0$ for all $i,j\in I$. For $q\in[0,1]$, define
\begin{equation}
s=\sum_{i\in I}g_i,\qquad S=\|s\|,\qquad
v_q=\sum_{i\in I}\|g_i\|^{q-1}g_i,\qquad
b_q=S\frac{v_q}{\|v_q\|}.
\label{eq:multi-reference}
\end{equation}
For nonempty $I$, both $S$ and $\|v_q\|$ are positive: the squared norms contain positive diagonal terms and nonnegative cross terms. Moreover,
\begin{equation}
s^\top v_q=\sum_{i\in I}\|g_i\|^{q+1}
+\sum_{\substack{i,j\in I\\i\ne j}}\|g_j\|^{q-1}g_i^\top g_j>0.
\end{equation}
Thus $s^\top b_q>0$. Given any mixing weight $\alpha\in[0,1]$, the vector $z=(1-\alpha)s+\alpha b_q$ is nonzero, and the same spherical compromise has the unique solution
\begin{equation}
\arg\min_{\|g\|=S}\bigl[(1-\alpha)\|g-s\|^2+\alpha\|g-b_q\|^2\bigr]
=S\frac{z}{\|z\|}.
\end{equation}
Expanding the objective gives a constant minus $2g^\top z$, so the result follows by maximizing the inner product on the sphere. This extends Proposition~\ref{prop:sphere} to the reference in Equation~\eqref{eq:multi-reference}. Zero gradients are excluded before evaluating the norm powers; if all gradients are zero, the update is zero.

This construction specifies compatible coordination given a mixing weight. A complete rule for multiple objectives additionally requires a choice of $\alpha$ from their joint geometry and a conflict operator for mixed relationships and task priorities. The implemented and evaluated rule in this paper specifies these choices for two objectives.

\section{Implementation and computational cost}
\subsection{Objective construction and implementation}
\label{app:implementation}
In the helpfulness--safety setting, the reward scores are calibrated using fixed statistics estimated from the held-out training-calibration subset. Each objective is centered within a response group. A common scale is then applied to the components, retaining their separate values. No evaluation prompt is part of the calibration subset.

For mathematical reasoning, let $C$ denote the correct responses in a group. The secondary advantage is
\[
A_{L,j}=\begin{cases}
r_{L,j}-|C|^{-1}\sum_{k\in C}r_{L,k},&j\in C,\ |C|>0,\\
0,&\text{otherwise}.
\end{cases}
\]
If all responses are incorrect, the secondary objective is inactive. If the length reward is constant on the correct subset, it contributes zero. The correctness advantage retains its primary GRPO normalization. For each response group, correctness scores are centered and divided by their sample standard deviation plus $10^{-6}$. In helpfulness--safety, let $B_i$ be the group-centered component broadcast over valid response tokens and $M$ their mask. The common normalization is
\[
A_i=\frac{B_i-\operatorname{Mean}_M(B_i)}{\sqrt{\operatorname{Var}_M(B_1+B_2)+10^{-8}}}\,M.
\]
Here $\operatorname{Var}_M$ is the masked variance used by the policy-training implementation. The masked variance applies Bessel's correction over valid response tokens.

The policy-loss adapter differentiates each reward-specific loss over all trainable policy parameters. Distributed reductions produce the global Gram entries. The shared regularization contribution is differentiated separately and included once. The maximized policy surrogate uses asymmetric clipping and a negative-advantage safeguard:
\[
\phi_{\mathrm{base}}(\rho,A)=\begin{cases}
\max\{\phi_{\mathrm{clip}}(\rho,A),\kappa A\},&A<0,\\
\phi_{\mathrm{clip}}(\rho,A),&A\geq0.
\end{cases}
\]
Both scenarios use $\kappa=3$, $\epsilon_-=0.2$, and $\epsilon_+=0.28$. Helpfulness--safety averages over all valid response tokens in the minibatch. Mathematics first averages valid tokens within each response and then averages responses. Exact-zero objective advantages remain zero after normalization, and an inactive objective contributes no gradient.

The mathematical setting uses the MSE log-ratio regularizer
\[
K(\theta;\mathcal B)=\frac{1}{|\mathcal B|}\sum_{j\in\mathcal B}\frac{1}{T_j}\sum_{t=1}^{T_j}\frac12\left(\log\pi_\theta(y_t^{(j)}\mid x_j,y_{<t}^{(j)})-\log\pi_{\mathrm{ref}}(y_t^{(j)}\mid x_j,y_{<t}^{(j)})\right)^2,
\]
where $T_j$ counts valid response tokens and $\pi_{\mathrm{ref}}$ is the frozen initial policy. The reduction is the same sequence mean of token means used for the policy objectives, and $\beta=0.0005$. Helpfulness--safety disables this regularizer. Both settings use zero entropy coefficient.

Table~\ref{tab:training-config} lists the training settings for the complete method in both scenarios. All policy parameters are trained. The initial policy is Qwen3-4B-Instruct-2507; helpfulness and harmlessness are scored by the Artessay Qwen2.5-7B-SafeRLHF reward and cost models.
\input{paper/tables/training_config}

Mathematical training uses eight H200 GPUs on one node. LoRA is disabled. Repeated-run aggregation is specified below.

\subsection{Reconciliation coefficients and implementation details}
\label{app:algorithm}
Algorithm~\ref{alg:orpg} computes the joint direction directly. Its rules can also be written as the weighted gradient in Equation~\eqref{eq:weighted-gradient}. All quantities below are evaluated at the current minibatch and parameters; their step superscripts are omitted.
For nonzero compatible gradients,
\[
\omega_i=\frac{S}{\|z\|_2}\left[(1-\alpha)+\frac{\alpha S n_i^{q-1}}{\|v_q\|_2}\right],\qquad i\in\{1,2\}.
\]
For a symmetric conflicting pair with $d=g_1^\top g_2<0$,
\[
\omega_i=1-\frac{d}{n_i^2},\qquad i\in\{1,2\}.
\]
For primary--secondary conflict resolution,
\[
\omega_p=1-\frac{d}{\|g_p\|_2^2},\qquad\omega_a=1.
\]
When either gradient is zero, $\omega_1=\omega_2=1$ gives the direct sum. These coefficients are algebraic expansions of the reconciliation rules. They are recomputed at each optimization minibatch and held fixed in the local surrogate's differentiation. The coefficient ratio $\omega_1/\omega_2$ weights the original gradients; the ratio $a_1/a_2$ in Appendix~\ref{app:proof} weights their unit directions.

\subsection{Training cost and reconciliation operations}
\label{app:cost}
The default mathematical run uses eight H200 GPUs for 100 optimizer steps. Summing the recorded duration of these steps gives 21.91 hours, or 175.31 GPU-hours, with a mean of 788.9 seconds per step. This measures accumulated training-step time, excluding queueing, intervals between training processes, initialization outside the step timers, discarded progress, and separate validation and evaluation. Table~\ref{tab:training-cost} reports the recorded components. Rollout generation averages 211.1 seconds per step and actor updates 474.4 seconds. Component timers may nest, so their entries are not an additive partition of the total.
\input{paper/tables/training_cost}

ORPG first obtains the full gradient of each policy objective. For two gradients in $\mathbb R^P$, reconciliation then uses the three independent Gram quantities $a=\|g_1\|^2$, $b=\|g_2\|^2$, and $d=g_1^\top g_2$. Computing these quantities and forming the final vector combination each take $O(P)$ arithmetic; the coefficient calculation is $O(1)$. With evenly distributed parameter shards over $D$ devices, the local vector operations take $O(P/D)$. The compatible coefficients can be computed from the Gram quantities using
\begin{equation}
V=\|v_q\|,\qquad t_i=(1-\alpha)+\alpha\frac{S}{V}n_i^{q-1},\qquad
\omega_i=\frac{St_i}{\sqrt{t_1^2a+t_2^2b+2t_1t_2d}},
\end{equation}
which gives $\mathcal R_+=\omega_1g_1+\omega_2g_2$. No parameter-space matrix or differentiation through the coefficients is needed. The distributed geometry reduction aggregates the three Gram quantities; full-gradient acquisition, FSDP synchronization, and the optimizer perform their own computation and communication. Storing the two objective gradients uses $O(P)$ memory. The recorded actor-update timer covers the policy update as a whole, including gradient acquisition and reconciliation.

\section{Evaluation protocols}
\subsection{Prompt construction and evaluation details}
\label{app:prompts}
\paragraph{Dataset sizes and calibration.}
Table~\ref{tab:dataset-sizes} lists the training, calibration, and evaluation splits. The three Alpaca subsets are disjoint. Mathematical evaluation contains 6,060 problems in total and generates four responses per problem.
\begin{table}[h]
\caption{Dataset sizes. Mathematical training uses DeepScaleR preview prompts.}
\label{tab:dataset-sizes}
\begin{center}
\small
\begin{tabular}{llr}
\toprule
Dataset & Use & Count\\\midrule
Alpaca & Training & 50,978\\
Alpaca & Reward calibration & 512\\
Alpaca & Evaluation & 512\\
HH-RLHF & Evaluation & 8,520\\
PKU-SafeRLHF & Evaluation & 8,211\\
AIME-24 & Evaluation & 30\\
AMC-22-23 & Evaluation & 83\\
MATH & Evaluation & 5,000\\
Minerva-Math & Evaluation & 272\\
OlympiadBench & Evaluation & 675\\\bottomrule
\end{tabular}
\end{center}
\end{table}

\paragraph{Mathematical training prompts.}
Each DeepScaleR training example supplies one user message. The exact content construction is:
\begin{quote}\ttfamily\small
\{problem\}\\
Please reason step by step, and put your final answer within \textbackslash boxed\{\}.
\end{quote}
The problem text is stripped of leading and trailing whitespace before appending the instruction. The reference answer is stored separately for reward evaluation and is not part of the user message.

\paragraph{Helpfulness--safety messages.}
The policy receives the dataset-provided user/assistant message sequence through its native tokenizer chat template, with \texttt{add\_generation\_prompt=True}. Evaluation preserves the final user request. When a prompt exceeds 512 tokens after template application, the adapter first removes the oldest complete conversation turns; if the remaining user message is still too long, it retains a token suffix that fits the prompt budget. This maintains a valid user-started, user-ended conversation.

\paragraph{Helpfulness--safety decoding.}
Evaluation uses one response per prompt, temperature 0.7, top-$p=1.0$, and a maximum of 1024 generated tokens. Reward scoring uses a maximum sequence length of 2048. All three datasets are evaluated in full for each seed. Their means are computed separately and then averaged with equal dataset weights. The policy generation and the two reward-model evaluations use their respective tokenizer interfaces.

\paragraph{Repeated-run aggregation.}
For each metric, let $m_k$ denote a complete run's result. We report $\bar m=N^{-1}\sum_km_k$ and sample standard deviation $s=\sqrt{\sum_k(m_k-\bar m)^2/(N-1)}$. A macro result is constructed within each run before computing its standard deviation. Results in both settings use three runs with seeds 42, 43, and 44.

\paragraph{Budget-level example.}
The initial policy's recorded AIME-24 evaluation illustrates the role of the shorter budgets in HV. At 2048, 4096, and 8192 tokens, accuracy is 18.33\%, 33.33\%, and 57.50\%, with mean response lengths of 1971, 3571, and 5761 tokens. The three-point HV is 0.2955, compared with an area of 0.1706 for the 8192 point alone. The additional points measure answer quality available at lower realized costs.

\subsection{Mathematical evaluation and hypervolume}
\label{app:metrics}
For dataset $D$, each prompt has four sampled responses. At budget $b$, the accuracy and mean length are
\[
a_{D,b}=\frac{1}{4|D|}\sum_{x\in D}\sum_{k=1}^4\mathbf{1}\{\text{response }(x,k)\text{ is correct at }b\},\qquad
\ell_{D,b}=\frac{1}{4|D|}\sum_{x\in D}\sum_{k=1}^4 L_{x,k,b}.
\]
All responses remain in the denominator, including unparseable answers. The shorter-budget views use exact token prefixes of the same generated responses. With $e_{D,b}=1-\min(1,\max(0,\ell_{D,b}/8192))$, define
\[
\operatorname{HV}_D=\operatorname{Area}\left(\bigcup_{b\in\{2048,4096,8192\}}
[0,e_{D,b}]\times[0,a_{D,b}]\right).
\]
The main table reports $a_{D,8192}$ and $\operatorname{HV}_D$. Avg is the equal-weight mean over the five datasets. Each repetition is summarized before calculating its mean and sample standard deviation; dataset standard deviations are not averaged to obtain a macro standard deviation.

Table~\ref{tab:math-budgets} provides the budget-specific macro accuracy and length measurements used to compute HV. The primary accuracy comparison uses the 8192-token budget; HV summarizes the union area defined above.
\input{paper/tables/math_budgets}

\section{Component comparisons and parameter sensitivity}
\subsection{Definitions of objective-wise comparison rules}
\label{app:variants}
All rules below act on the separate reward-specific policy gradients $g_1,g_2$ before the shared regularization contribution. Sum uses $g_1+g_2$ for every pair. PCGrad uses the symmetric conflict projection in Algorithm~\ref{alg:orpg} when $g_1^\top g_2<0$ and the direct sum otherwise. Thus, PCGrad is the compatible-coordination-off variant in the HS setting. The conflict-resolution-off variant uses Equation~\eqref{eq:compatible} for compatible gradients and the direct sum for conflicting gradients. Removing both components gives Sum. The CAGrad and Aligned-MTL rows use their respective joint-gradient constructions within the same objective-wise policy interface \citep{liu2021cagrad,senushkin2023aligned}.

\input{paper/sections/math_components}

\input{paper/sections/supplement}

\section{Training dynamics and measurements}
\subsection{Additional training-gradient analysis}
\label{app:telemetry}
The measurements cover all 100 optimizer steps for the four component versions. Figure~\ref{fig:training-signals} presents gradient cosine and Useful RMS in the main text. Table~\ref{tab:phases} summarizes these signals together with conflict and projection rates over the same three training intervals.

\begin{figure}[h]
\centering
\includegraphics[width=0.86\linewidth]{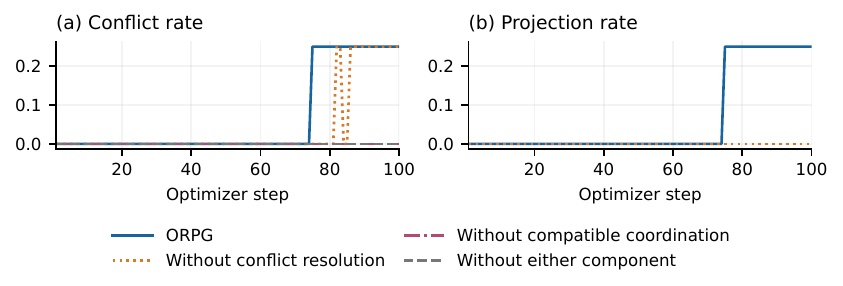}
\caption{Unsmoothed conflict and projection statistics. (a) Conflict rates for all four component versions; the two zero-conflict curves overlap. (b) Projection rates for ORPG and Without conflict resolution.}
\label{fig:conflict-details}
\end{figure}

Figure~\ref{fig:conflict-details}(a) shows when conflicting gradient pairs occur. The two versions retaining compatible coordination encounter conflicts later in training, while the other two record zero conflicts. Figure~\ref{fig:conflict-details}(b) shows that ORPG projects those pairs, whereas Without conflict resolution leaves them unprojected. A positive step-average cosine can coexist with conflicts on individual updates within that step.

\input{paper/tables/training_phases}

\begin{figure}[h]
\centering
\includegraphics[width=0.62\linewidth]{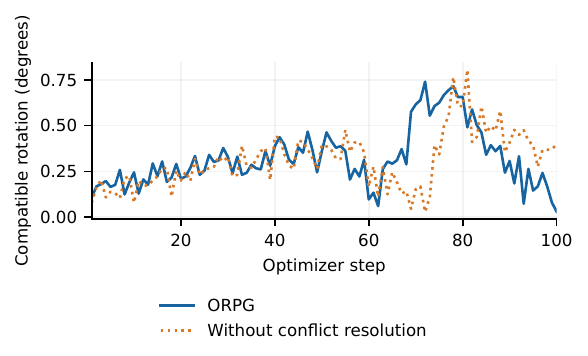}
\caption{Compatible-branch rotation for the two versions retaining compatible coordination. Values are unsmoothed step aggregates.}
\label{fig:compatible-rotation}
\end{figure}

Figure~\ref{fig:compatible-rotation} supplements these relationships with the angle between the original sum and the compatible output. The statistic uses their normalized vector difference; conflict-branch calls contribute zero. The plot compares the two versions that apply compatible coordination.

\subsection{Training reward and advantage measurements}
\label{app:reward-dynamics}
Training rewards are the calibrated Useful and Harmless values computed on the rollout batch. Evaluation reports raw reward-model scores on held-out prompts. The two quantities share objective meanings but have different scales and samples. For display, all reward curves average rewards across training runs at each step and then apply a trailing five-step mean. No variability band is inferred from temporal smoothing.

The component advantage RMS is the square root of the mean squared advantage over valid response tokens, recorded separately for each objective. All four component versions use shared-scale normalization and exact-zero handling. The values in Table~\ref{tab:training-final} average steps 81--100 within each trajectory and then average actual training runs. The Useful RMS measurement is interpreted together with the observed Useful reward growth.
\begin{table}[h]
\caption{Late-training calibrated rewards and Useful advantage RMS (steps 81--100).}
\label{tab:training-final}
\begin{center}
\small
\begin{tabular}{lrrr}
\toprule
Method & Useful & Harmless & Useful RMS \\
\midrule
GDPO & 1.1568 & 1.4582 & -- \\
GD$^2$PO & 1.1577 & 1.4620 & -- \\
Without either component & 1.1532 & 1.7368 & 0.5207 \\
Without compatible coordination & 1.1738 & 1.7676 & 0.5278 \\
Without conflict resolution & 1.4924 & 1.9990 & 0.8600 \\
ORPG & \textbf{1.5217} & \textbf{2.0051} & 0.8868 \\
\bottomrule
\end{tabular}
\end{center}
\end{table}

%% file: paper/tables/training_config.tex
\begin{table}[!htbp]
\caption{Training settings. Token limits refer to training; evaluation budgets are specified separately.}
\label{tab:training-config}
\begin{center}
\small
\setlength{\tabcolsep}{3pt}
\begin{tabular}{p{0.32\linewidth}p{0.27\linewidth}p{0.34\linewidth}}
\toprule
Parameter & Helpfulness--safety & Correctness--cost \\
\midrule
Training steps & 100 & 100 \\
Batch / minibatch / rollout group & 512 / 128 / 4 & 512 / 64 / 8 \\
PPO epochs & 1 & 1 \\
Learning rate & $2\times10^{-6}$ & $10^{-6}$ \\
Optimizer & AdamW & AdamW \\
Schedule / warmup & Constant / none & Constant / none \\
Adam $(\beta_1,\beta_2)$ / $\epsilon$ & $(0.9,0.999)$ / $10^{-8}$ & $(0.9,0.999)$ / $10^{-8}$ \\
Weight decay & 0.01 & 0.01 \\
Prompt / response limit & 512 / 1024 & 1024 / 8000 \\
Temperature / top-$p$ / top-$k$ & 0.7 / 1.0 / $-1$ & 1.0 / 1.0 / $-1$ \\
PPO clip lower / upper & 0.2 / 0.28 & 0.2 / 0.28 \\
Loss reduction & Token mean & Sequence mean of token means \\
Gradient norm clip & 1.0 & 1.0 \\
KL coefficient / type & 0 / disabled & 0.0005 / MSE \\
Entropy coefficient & 0 & 0 \\
Compatible $q$ / $\lambda$ & 0.5 / 0.25 & 0.5 / 0.25 \\
Preserve sum norm & Yes & Yes \\
Conflict rule & Symmetric & Correctness priority \\
Reward weights & 1 / 1 & 1 / 1 \\
Length threshold & --- & $\tau=4000$ \\
Advantage construction & Group centered, shared scale & Primary GRPO; correct-subset length \\
Precision / sharding & bfloat16 / FSDP & bfloat16 / FSDP \\
\bottomrule
\end{tabular}
\end{center}
\end{table}

%% file: paper/tables/training_cost.tex
\begin{table}[!ht]
\caption{Recorded training-step costs for the default mathematical run on eight H200 GPUs. Component timers can overlap or nest.}
\label{tab:training-cost}
\begin{center}
\small
\begin{tabular}{lrr}
\toprule
Recorded operation & Seconds per step & Total hours \\
\midrule
Total training step & 788.9 & 21.91 \\
Rollout generation & 211.1 & 5.86 \\
Actor update & 474.4 & 13.18 \\
Old-policy log probabilities & 48.6 & 1.35 \\
Reference-policy log probabilities & 46.4 & 1.29 \\
Advantage computation & 2.4 & 0.07 \\
Rollout weight update & 3.9 & 0.11 \\
\bottomrule
\end{tabular}
\end{center}
\end{table}

%% file: paper/tables/math_budgets.tex
\begin{table}[!htbp]
\caption{Budget-specific measurements used to compute hypervolume. Accuracy (Acc) is in percent; length (Len) is in tokens.}
\label{tab:math-budgets}
\begin{center}
\small
\setlength{\tabcolsep}{3pt}
\begin{tabular}{lrrrrrr}
\toprule
Method & \multicolumn{2}{c}{2048} & \multicolumn{2}{c}{4096} & \multicolumn{2}{c}{8192} \\
\cmidrule(lr){2-3}\cmidrule(lr){4-5}\cmidrule(lr){6-7}
 & Acc $\uparrow$ & Len $\downarrow$ & Acc $\uparrow$ & Len $\downarrow$ & Acc $\uparrow$ & Len $\downarrow$ \\
\midrule
Base & \result{43.76}{0.34} & \result{1434}{2} & \result{53.66}{0.48} & \result{2232}{1} & \result{65.52}{0.70} & \result{3165}{12} \\
GRPO & \result{51.09}{0.66} & \result{1299}{54} & \result{\textbf{60.40}}{0.48} & \result{1556}{70} & \result{60.69}{0.52} & \result{1568}{66} \\
GDPO & \result{51.94}{0.19} & \result{\textbf{1192}}{19} & \result{59.00}{0.41} & \result{\textbf{1385}}{36} & \result{59.35}{0.12} & \result{\textbf{1392}}{38} \\
GD$^2$PO & \result{\textbf{52.31}}{1.27} & \result{1204}{19} & \result{59.84}{1.03} & \result{1439}{33} & \result{60.26}{0.85} & \result{1456}{29} \\
ORPG & \result{47.04}{0.32} & \result{1406}{3} & \result{59.00}{0.27} & \result{2076}{8} & \result{\textbf{66.66}}{0.26} & \result{2554}{22} \\
\bottomrule
\end{tabular}
\end{center}
\end{table}

%% file: paper/sections/math_components.tex
\subsection{Mathematical component comparisons}
\label{app:math-components}
The four versions share the mathematical reward definitions and objective-wise advantage construction, including length advantages centered within the correct-response subset. Without compatible coordination retains correctness-priority conflict projection and directly sums compatible gradients. Without conflict resolution retains compatible coordination and sums conflicting gradients. Without either component sums the two objective gradients in every case. 

\begin{table}[!htbp]
\caption{Mathematical component comparisons. Accuracy and length use the 8192-token budget; HV aggregates the three budget points.}
\label{tab:math-components}
\begin{center}
\small
\begin{tabular}{lrrr}
\toprule
Update & Accuracy (\%) $\uparrow$ & Length $\downarrow$ & HV $\uparrow$ \\
\midrule
ORPG & \result{\textbf{66.66}}{0.26} & \result{2554.14}{21.83} & \result{0.5226}{0.0017} \\
Without compatible coordination & \result{66.15}{0.19} & \result{2447.87}{30.61} & \result{\textbf{0.5270}}{0.0025} \\
Without conflict resolution & \result{65.46}{0.55} & \result{2456.62}{18.64} & \result{0.5203}{0.0026} \\
Without either component & \result{66.04}{0.14} & \result{\textbf{2399.56}}{5.39} & \result{0.5236}{0.0021} \\
\bottomrule
\end{tabular}
\end{center}
\end{table}

Table~\ref{tab:math-components} shows that the complete rule achieves the highest full-budget accuracy. It exceeds the version without conflict resolution by 1.20 percentage points, connecting correctness-priority projection to improved answer quality when compatible coordination is retained. Its accuracy also exceeds the versions without compatible coordination and without either component by 0.51 and 0.62 percentage points. The full rule uses longer responses to attain this accuracy. The version without compatible coordination achieves the highest HV, while direct summation gives the shortest responses. These comparisons show how the components affect the accuracy--cost trade-off under the shared correctness-first objective construction.

%% file: paper/sections/supplement.tex
\subsection{Sensitivity to compatible coordination parameters}
\label{app:sensitivity}
We vary the reference exponent $q$ and maximum mixing strength $\lambda$ individually around the shared default $(q,\lambda)=(0.5,0.25)$. The $q$ scan uses $\{0.25,0.5,0.75\}$ at $\lambda=0.25$; the $\lambda$ scan uses $\{0.125,0.25,0.5\}$ at $q=0.5$. The scans share their default point, giving five configurations per setting. Scores follow the main evaluation protocol: full-set mean@1 on 17,243 HS prompts, and four samples per problem on 6,060 mathematical problems with exact-prefix budgets of 2048, 4096, and 8192 tokens. Means and sample standard deviations are computed across the three runs after dataset-level macro aggregation.

\input{paper/tables/sensitivity_hs}
\input{paper/tables/sensitivity_math}

Tables~\ref{tab:sensitivity-hs} and~\ref{tab:sensitivity-math} show that the main gains persist across the tested parameter range. Useful ranges from 5.482 to 5.589 and Harmless from 6.806 to 6.904, exceeding the strongest external baseline on each metric in Table~\ref{tab:hs}. Mathematical full-budget accuracy ranges from 65.43\% to 66.66\% and HV from 0.5205 to 0.5243; all five configurations exceed the external training baselines in Tables~\ref{tab:math} and~\ref{tab:math-hv} on both metrics.

\begin{figure}[!ht]
\centering
\includegraphics[width=0.95\linewidth]{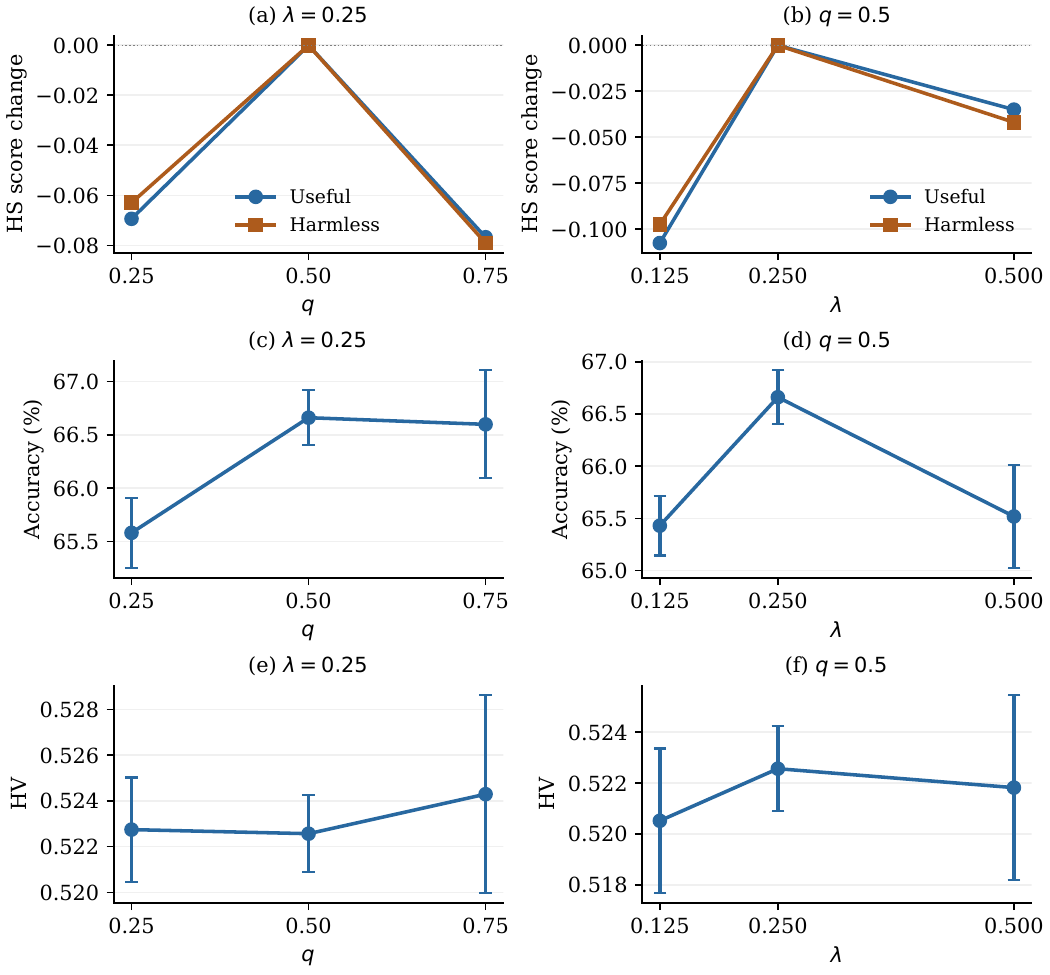}
\caption{Parameter sensitivity with one parameter varied at a time. (a--b) HS score changes relative to the default; raw scores appear in Table~\ref{tab:sensitivity-hs}. (c--d) Full-budget mathematical accuracy. (e--f) Three-budget mathematical HV.}
\label{fig:sensitivity}
\end{figure}

The default achieves the highest HS scores and full-budget mathematical accuracy among these configurations. Its exponent $q=0.5$ changes the reference amplitude ratio to the square root of the original norm ratio, retaining magnitude information while moderating large-norm contributions. The shared $\lambda=0.25$ provides a moderate coordination strength supported by both task settings. Figure~\ref{fig:sensitivity} shows the local response to each parameter. In mathematics, $q=0.75$ gives the highest three-budget HV, while the default gives the highest full-budget accuracy. Table~\ref{tab:sensitivity-budgets} further shows that the accuracy ranking changes with the token budget: $q=0.25$ leads at 2048 tokens, $\lambda=0.125$ at 4096, and the default at 8192. These outcomes describe how the parameters adjust the accuracy--cost trade-off across budgets.

\input{paper/tables/sensitivity_budgets}

%% file: paper/tables/sensitivity_hs.tex
\begin{table}[!ht]
\caption{Sensitivity on helpfulness--safety. Scores are macro averages over the three evaluation sets.}
\label{tab:sensitivity-hs}
\begin{center}
\small
\begin{tabular}{rrrr}
\toprule
$q$ & $\lambda$ & Useful $\uparrow$ & Harmless $\uparrow$ \\
\midrule
0.25 & 0.25 & \result{5.5197}{0.0005} & \result{6.8409}{0.0012} \\
0.5 & 0.25 & \result{\textbf{5.5891}}{0.0004} & \result{\textbf{6.9038}}{0.0014} \\
0.75 & 0.25 & \result{5.5124}{0.0004} & \result{6.8248}{0.0011} \\
0.5 & 0.125 & \result{5.4816}{0.0017} & \result{6.8064}{0.0015} \\
0.5 & 0.5 & \result{5.5542}{0.0012} & \result{6.8620}{0.0009} \\
\bottomrule
\end{tabular}
\end{center}
\end{table}

%% file: paper/tables/sensitivity_math.tex
\begin{table}[!ht]
\caption{Sensitivity on mathematics: full-budget accuracy and length, and three-budget hypervolume.}
\label{tab:sensitivity-math}
\begin{center}
\small
\begin{tabular}{rrrrr}
\toprule
$q$ & $\lambda$ & Accuracy (\%) $\uparrow$ & Length $\downarrow$ & HV $\uparrow$ \\
\midrule
0.25 & 0.25 & \result{65.58}{0.33} & \result{2425.42}{38.69} & \result{0.5227}{0.0023} \\
0.5 & 0.25 & \result{\textbf{66.66}}{0.26} & \result{2554.14}{21.83} & \result{0.5226}{0.0017} \\
0.75 & 0.25 & \result{66.60}{0.51} & \result{2613.33}{12.82} & \result{\textbf{0.5243}}{0.0043} \\
0.5 & 0.125 & \result{65.43}{0.29} & \result{\textbf{2413.83}}{14.08} & \result{0.5205}{0.0028} \\
0.5 & 0.5 & \result{65.52}{0.49} & \result{2434.32}{11.65} & \result{0.5218}{0.0036} \\
\bottomrule
\end{tabular}
\end{center}
\end{table}

%% file: paper/tables/sensitivity_budgets.tex
\begin{table}[!ht]
\caption{Mathematical sensitivity at each token budget. Accuracy is in percent and length in tokens.}
\label{tab:sensitivity-budgets}
\begin{center}
\small
\setlength{\tabcolsep}{5.8pt}
\begin{tabular}{rrrrrrrr}
\toprule
 & & \multicolumn{2}{c}{2048} & \multicolumn{2}{c}{4096} & \multicolumn{2}{c}{8192} \\ \cmidrule(lr){3-4}\cmidrule(lr){5-6}\cmidrule(lr){7-8} $q$ & $\lambda$ & Acc. $\uparrow$ & Length $\downarrow$ & Acc. $\uparrow$ & Length $\downarrow$ & Acc. $\uparrow$ & Length $\downarrow$ \\
\midrule
0.25 & 0.25 & \result{\textbf{47.77}}{0.40} & \result{\textbf{1374.0}}{1.7} & \result{59.31}{0.44} & \result{\textbf{2015.2}}{11.0} & \result{65.58}{0.33} & \result{2425.4}{38.7} \\
0.5 & 0.25 & \result{47.04}{0.32} & \result{1406.3}{3.2} & \result{59.00}{0.27} & \result{2075.9}{7.9} & \result{\textbf{66.66}}{0.26} & \result{2554.1}{21.8} \\
0.75 & 0.25 & \result{47.26}{0.61} & \result{1390.6}{3.2} & \result{59.33}{0.36} & \result{2079.0}{19.7} & \result{66.60}{0.51} & \result{2613.3}{12.8} \\
0.5 & 0.125 & \result{47.34}{0.49} & \result{1408.5}{3.1} & \result{\textbf{60.55}}{0.77} & \result{2035.7}{14.5} & \result{65.43}{0.29} & \result{\textbf{2413.8}}{14.1} \\
0.5 & 0.5 & \result{47.27}{0.44} & \result{1380.7}{2.0} & \result{59.44}{0.21} & \result{2031.8}{2.9} & \result{65.52}{0.49} & \result{2434.3}{11.6} \\
\bottomrule
\end{tabular}
\end{center}
\end{table}

%% file: paper/tables/training_phases.tex
\begin{table}[h]
\caption{Training-stage gradient relationships and Useful advantage RMS. All versions use the same three step intervals.}
\label{tab:phases}
\begin{center}
\fontsize{8}{10}\selectfont
\begin{tabular}{llrrrr}
\toprule
Update & Steps & Cosine & Conflict & Projection & Useful RMS \\
\midrule
\multirow{3}{*}{ORPG} & 1--33 & 0.661 & 0.000 & 0.000 & 0.490 \\
 & 34--66 & 0.584 & 0.000 & 0.000 & 0.500 \\
 & 67--100 & 0.265 & 0.191 & 0.191 & 0.857 \\
\multirow{3}{*}{Without compatible coordination} & 1--33 & 0.690 & 0.000 & 0.000 & 0.492 \\
 & 34--66 & 0.616 & 0.000 & 0.000 & 0.488 \\
 & 67--100 & 0.522 & 0.000 & 0.000 & 0.507 \\
\multirow{3}{*}{Without conflict resolution} & 1--33 & 0.661 & 0.000 & 0.000 & 0.493 \\
 & 34--66 & 0.576 & 0.000 & 0.000 & 0.484 \\
 & 67--100 & 0.398 & 0.125 & 0.000 & 0.759 \\
\multirow{3}{*}{Without either component} & 1--33 & 0.692 & 0.000 & 0.000 & 0.493 \\
 & 34--66 & 0.619 & 0.000 & 0.000 & 0.485 \\
 & 67--100 & 0.523 & 0.000 & 0.000 & 0.504 \\
\bottomrule
\end{tabular}
\end{center}
\end{table}

%% file: paper/references.bib
@misc{shao2024deepseekmath,
      title={{DeepSeekMath: Pushing the Limits of Mathematical Reasoning in Open Language Models}},
      author={Zhihong Shao and Peiyi Wang and Qihao Zhu and Runxin Xu and Junxiao Song and Xiao Bi and Haowei Zhang and Mingchuan Zhang and Y. K. Li and Y. Wu and Daya Guo},
      year={2024},
      eprint={2402.03300},
      archivePrefix={arXiv},
      primaryClass={cs.CL},
      url={https://arxiv.org/abs/2402.03300},
}

@misc{liu2026gdpo,
      title={{GDPO: Group reward-Decoupled Normalization Policy Optimization for Multi-reward RL Optimization}},
      author={Shih-Yang Liu and Xin Dong and Ximing Lu and Shizhe Diao and Peter Belcak and Mingjie Liu and Min-Hung Chen and Hongxu Yin and Yu-Chiang Frank Wang and Kwang-Ting Cheng and Yejin Choi and Jan Kautz and Pavlo Molchanov},
      year={2026},
      eprint={2601.05242},
      archivePrefix={arXiv},
      primaryClass={cs.CL},
      url={https://arxiv.org/abs/2601.05242},
}

@misc{liu2026gd2po,
      title={{GD$^2$PO: Mitigating Multi-Reward Conflicts via Group-Dynamic reward-Decoupled Policy Optimization}},
      author={Haotian Liu and Yihao Liu and Jingwei Ni and Siyuan Huang and Xinpeng Liu and Pengyu Cheng and Jiajun Song and Ruijin Ding and Junfeng Li and Zhechao Yu and Mengyu Zhou and Hongteng Xu and Xiaoxi Jiang and Guanjun Jiang},
      year={2026},
      eprint={2606.16771},
      archivePrefix={arXiv},
      primaryClass={cs.LG},
      url={https://arxiv.org/abs/2606.16771},
}

@inproceedings{yu2020pcgrad,
 author = {Yu, Tianhe and Kumar, Saurabh and Gupta, Abhishek and Levine, Sergey and Hausman, Karol and Finn, Chelsea},
 booktitle = {{Advances in Neural Information Processing Systems}},
 editor = {H. Larochelle and M. Ranzato and R. Hadsell and M.F. Balcan and H. Lin},
 pages = {5824--5836},
 publisher = {Curran Associates, Inc.},
 title = {{Gradient Surgery for Multi-Task Learning}},
 url = {https://proceedings.neurips.cc/paper_files/paper/2020/file/3fe78a8acf5fda99de95303940a2420c-Paper.pdf},
 volume = {33},
 year = {2020}
}

@inproceedings{liu2021cagrad,
 author = {Liu, Bo and Liu, Xingchao and Jin, Xiaojie and Stone, Peter and Liu, Qiang},
 booktitle = {{Advances in Neural Information Processing Systems}},
 editor = {M. Ranzato and A. Beygelzimer and Y. Dauphin and P.S. Liang and J. Wortman Vaughan},
 pages = {18878--18890},
 publisher = {Curran Associates, Inc.},
 title = {{Conflict-Averse Gradient Descent for Multi-task learning}},
 url = {https://proceedings.neurips.cc/paper_files/paper/2021/file/9d27fdf2477ffbff837d73ef7ae23db9-Paper.pdf},
 volume = {34},
 year = {2021}
}

@InProceedings{senushkin2023aligned,
    author    = {Senushkin, Dmitry and Patakin, Nikolay and Kuznetsov, Arseny and Konushin, Anton},
    title     = {{Independent Component Alignment for Multi-Task Learning}},
    booktitle = {{Proceedings of the IEEE/CVF Conference on Computer Vision and Pattern Recognition (CVPR)}},
    month     = {June},
    year      = {2023},
    pages     = {20083-20093}
}

@inproceedings{wang2021gradvac,
  author = {Wang, Zirui and Tsvetkov, Yulia and Firat, Orhan and Cao, Yuan},
  title = {{Gradient Vaccine: Investigating and Improving Multi-task Optimization in Massively Multilingual Models}},
  booktitle = {International Conference on Learning Representations},
  year = {2021},
  url = {https://openreview.net/pdf/4958372042631716242a4b3f1a10231548614522.pdf}
}

@misc{ichihara2025mogrpo,
 title = {{MO-GRPO: Mitigating Reward Hacking of Group Relative Policy Optimization on Multi-Objective Problems}},
 author = {Yuki Ichihara and Yuu Jinnai and Tetsuro Morimura and Mitsuki Sakamoto and Ryota Mitsuhashi and Eiji Uchibe},
 year = {2025},
 eprint = {2509.22047},
 archivePrefix = {arXiv},
 url = {https://arxiv.org/abs/2509.22047}
}

@misc{he2025pama,
 title = {{Pareto Multi-Objective Alignment for Language Models}},
 author = {Qiang He and Setareh Maghsudi},
 year = {2025},
 eprint = {2508.07768},
 archivePrefix = {arXiv},
 url = {https://arxiv.org/abs/2508.07768}
}

@misc{lu2025dynamicreward,
 title = {{Learning to Optimize Multi-Objective Alignment Through Dynamic Reward Weighting}},
 author = {Yining Lu and Zilong Wang and Shiyang Li and Xin Liu and Changlong Yu and Qingyu Yin and Zhan Shi and Zixuan Zhang and Meng Jiang},
 year = {2025},
 eprint = {2509.11452},
 archivePrefix = {arXiv},
 url = {https://arxiv.org/abs/2509.11452}
}

@misc{pavlenko2026blockwise,
 title = {{Blockwise Advantage Estimation for Multi-Objective RL with Verifiable Rewards}},
 author = {Kirill Pavlenko and Alexander Golubev and Simon Karasik and Boris Yangel},
 year = {2026},
 eprint = {2602.10231},
 archivePrefix = {arXiv},
 url = {https://arxiv.org/abs/2602.10231}
}

@misc{schulman2017proximalpolicyoptimizationalgorithms,
      title={Proximal Policy Optimization Algorithms},
      author={John Schulman and Filip Wolski and Prafulla Dhariwal and Alec Radford and Oleg Klimov},
      year={2017},
      eprint={1707.06347},
      archivePrefix={arXiv},
      primaryClass={cs.LG},
      url={https://arxiv.org/abs/1707.06347},
}

@inproceedings{li-etal-2025-gradient,
    title = "Gradient-Adaptive Policy Optimization: Towards Multi-Objective Alignment of Large Language Models",
    author = "Li, Chengao  and
      Zhang, Hanyu  and
      Xu, Yunkun  and
      Xue, Hongyan  and
      Ao, Xiang  and
      He, Qing",
    editor = "Che, Wanxiang  and
      Nabende, Joyce  and
      Shutova, Ekaterina  and
      Pilehvar, Mohammad Taher",
    booktitle = "Proceedings of the 63rd Annual Meeting of the Association for Computational Linguistics (Volume 1: Long Papers)",
    month = jul,
    year = "2025",
    address = "Vienna, Austria",
    publisher = "Association for Computational Linguistics",
    url = "https://aclanthology.org/2025.acl-long.549/",
    doi = "10.18653/v1/2025.acl-long.549",
    pages = "11214--11232",
    ISBN = "979-8-89176-251-0"
}

@inproceedings{chen2018gradnorm,
 title={{GradNorm}: Gradient Normalization for Adaptive Loss Balancing in Deep Multitask Networks},
 author={Chen, Zhao and Badrinarayanan, Vijay and Lee, Chen-Yu and Rabinovich, Andrew},
 booktitle={Proceedings of the 35th International Conference on Machine Learning},
 year={2018}, volume={80}, pages={794--803}, series={Proceedings of Machine Learning Research}, publisher={PMLR},
 url={https://proceedings.mlr.press/v80/chen18a.html}
}

@inproceedings{sener2018multi,
 title={Multi-Task Learning as Multi-Objective Optimization},
 author={Sener, Ozan and Koltun, Vladlen},
 booktitle={Advances in Neural Information Processing Systems},
 year={2018}, volume={31},
 url={https://proceedings.neurips.cc/paper_files/paper/2018/hash/432aca3a1e345e339f35a30c8f65edce-Abstract.html}
}

@misc{du2018auxiliary,
 title={Adapting Auxiliary Losses Using Gradient Similarity},
 author={Du, Yunshu and Czarnecki, Wojciech M. and Jayakumar, Siddhant M. and Pascanu, Razvan and Lakshminarayanan, Balaji},
 year={2018}, eprint={1812.02224}, archivePrefix={arXiv}, url={https://arxiv.org/abs/1812.02224}
}

@article{ouyang2022instructgpt,
  title = {Training language models to follow instructions with human feedback},
  author = {Long Ouyang and Jeff Wu and Xu Jiang and Diogo Almeida and Carroll L. Wainwright and Pamela Mishkin and Chong Zhang and Sandhini Agarwal and Katarina Slama and Alex Ray and John Schulman and Jacob Hilton and Fraser Kelton and Luke Miller and Maddie Simens and Amanda Askell and Peter Welinder and Paul Christiano and Jan Leike and Ryan Lowe},
  year = {2022},
  journal = {arXiv preprint arXiv:2203.02155},
  eprint = {2203.02155},
  archivePrefix = {arXiv},
  url = {https://arxiv.org/abs/2203.02155}
}

@article{dai2023saferlhf,
  title = {Safe {RLHF}: Safe Reinforcement Learning from Human Feedback},
  author = {Josef Dai and Xuehai Pan and Ruiyang Sun and Jiaming Ji and Xinbo Xu and Mickel Liu and Yizhou Wang and Yaodong Yang},
  year = {2023},
  journal = {arXiv preprint arXiv:2310.12773},
  eprint = {2310.12773},
  archivePrefix = {arXiv},
  url = {https://arxiv.org/abs/2310.12773}
}

@misc{Li2026RethinkingTR,
  title={Rethinking the Role of Entropy in Optimizing Tool-Use Behaviors for Large Language Model Agents},
  author={Ze-Ping Li and Hongru Wang and Yiwen Zhao and Guanhua Chen and Yixia Li and Keyang Chen and Yixin Cao and Guangnan Ye and Hongfeng Chai and Zhen-Fei Yin},
  journal={ArXiv},
  year={2026},
  volume={abs/2602.02050},
  url={https://api.semanticscholar.org/CorpusID:285271471}
}

@misc{Zhao2026TowardsBA,
  title={Towards Better Agents for Multi-Turn User Interaction: The Next User Turn Is More Than Context},
  author={Yiwen Zhao and Zhihao Wen and Yuchen Mao and Mingxuan Jiang and Yihao Hu and Pan Wang and Xin Zhang and Wei Wu},
  year={2026},
  url={https://api.semanticscholar.org/CorpusID:291184908}
}

@article{aggarwal2025l1,
  title = {{L1}: Controlling How Long A Reasoning Model Thinks With Reinforcement Learning},
  author = {Pranjal Aggarwal and Sean Welleck},
  year = {2025},
  journal = {arXiv preprint arXiv:2503.04697},
  eprint = {2503.04697},
  archivePrefix = {arXiv},
  url = {https://arxiv.org/abs/2503.04697}
}

@article{luo2025o1pruner,
  title = {{O1-Pruner}: Length-Harmonizing Fine-Tuning for {O1}-Like Reasoning Pruning},
  author = {Haotian Luo and Li Shen and Haiying He and Yibo Wang and Shiwei Liu and Wei Li and Naiqiang Tan and Xiaochun Cao and Dacheng Tao},
  year = {2025},
  journal = {arXiv preprint arXiv:2501.12570},
  eprint = {2501.12570},
  archivePrefix = {arXiv},
  url = {https://arxiv.org/abs/2501.12570}
}

@article{yi2025shorterbetter,
  title = {{ShorterBetter}: Guiding Reasoning Models to Find Optimal Inference Length for Efficient Reasoning},
  author = {Jingyang Yi and Jiazheng Wang and Sida Li},
  year = {2025},
  journal = {arXiv preprint arXiv:2504.21370},
  eprint = {2504.21370},
  archivePrefix = {arXiv},
  url = {https://arxiv.org/abs/2504.21370}
}

@article{liu2025dler,
  title = {{DLER}: Doing Length pEnalty Right -- Incentivizing More Intelligence per Token via Reinforcement Learning},
  author = {Shih-Yang Liu and Xin Dong and Ximing Lu and Shizhe Diao and Mingjie Liu and Min-Hung Chen and Hongxu Yin and Yu-Chiang Frank Wang and Kwang-Ting Cheng and Yejin Choi and Jan Kautz and Pavlo Molchanov},
  year = {2025},
  journal = {arXiv preprint arXiv:2510.15110},
  eprint = {2510.15110},
  archivePrefix = {arXiv},
  url = {https://arxiv.org/abs/2510.15110}
}

@article{shrivastava2025gfpo,
  title = {Sample More to Think Less: Group Filtered Policy Optimization for Concise Reasoning},
  author = {Vaishnavi Shrivastava and Ahmed Awadallah and Vidhisha Balachandran and Shivam Garg and Harkirat Behl and Dimitris Papailiopoulos},
  year = {2025},
  journal = {arXiv preprint arXiv:2508.09726},
  eprint = {2508.09726},
  archivePrefix = {arXiv},
  url = {https://arxiv.org/abs/2508.09726}
}

@article{liu2025laser,
  title = {Learn to Reason Efficiently with Adaptive Length-based Reward Shaping},
  author = {Wei Liu and Ruochen Zhou and Yiyun Deng and Yuzhen Huang and Junteng Liu and Yuntian Deng and Yizhe Zhang and Junxian He},
  year = {2025},
  journal = {arXiv preprint arXiv:2505.15612},
  eprint = {2505.15612},
  archivePrefix = {arXiv},
  url = {https://arxiv.org/abs/2505.15612}
}

@article{sheng2024hybridflow,
  title = {{HybridFlow}: A Flexible and Efficient {RLHF} Framework},
  author = {Guangming Sheng and Chi Zhang and Zilingfeng Ye and Xibin Wu and Wang Zhang and Ru Zhang and Yanghua Peng and Haibin Lin and Chuan Wu},
  year = {2024},
  journal = {arXiv preprint arXiv:2409.19256},
  eprint = {2409.19256},
  archivePrefix = {arXiv},
  url = {https://arxiv.org/abs/2409.19256}
}

@misc{yang2025qwen3technicalreport,
      title={Qwen3 Technical Report},
      author={An Yang and Anfeng Li and Baosong Yang and Beichen Zhang and Binyuan Hui and Bo Zheng and Bowen Yu and Chang Gao and Chengen Huang and Chenxu Lv and Chujie Zheng and Dayiheng Liu and Fan Zhou and Fei Huang and Feng Hu and Hao Ge and Haoran Wei and Huan Lin and Jialong Tang and Jian Yang and Jianhong Tu and Jianwei Zhang and Jianxin Yang and Jiaxi Yang and Jing Zhou and Jingren Zhou and Junyang Lin and Kai Dang and Keqin Bao and Kexin Yang and Le Yu and Lianghao Deng and Mei Li and Mingfeng Xue and Mingze Li and Pei Zhang and Peng Wang and Qin Zhu and Rui Men and Ruize Gao and Shixuan Liu and Shuang Luo and Tianhao Li and Tianyi Tang and Wenbiao Yin and Xingzhang Ren and Xinyu Wang and Xinyu Zhang and Xuancheng Ren and Yang Fan and Yang Su and Yichang Zhang and Yinger Zhang and Yu Wan and Yuqiong Liu and Zekun Wang and Zeyu Cui and Zhenru Zhang and Zhipeng Zhou and Zihan Qiu},
      year={2025},
      eprint={2505.09388},
      archivePrefix={arXiv},
      primaryClass={cs.CL},
      url={https://arxiv.org/abs/2505.09388},
}

@misc{qwen2025qwen25technicalreport,
      title={Qwen2.5 Technical Report},
      author={Qwen An Yang and Baosong Yang and Beichen Zhang and Binyuan Hui and Bo Zheng and Bo-Wen Yu and Chengyuan Li and Dayiheng Liu and Fei Huang and Guanting Dong and Haoran Wei and Huan Lin and Jian Yang and Jianhong Tu and Jianwei Zhang and Jianxin Yang and Jiaxin Yang and Jingren Zhou and Junyang Lin and Kai Dang and Keming Lu and Keqin Bao and Kexin Yang and Le Yu and Mei Li and Mingfeng Xue and Pei Zhang and Qin Zhu and Rui Men and Runji Lin and Tianhao Li and Tingyu Xia and Xingzhang Ren and Xuancheng Ren and Yang Fan and Yang Su and Yi-Chao Zhang and Yunyang Wan and Yuqi Liu and Zeyu Cui and Zhen-Ru Zhang and Zihan Qiu and Shanghaoran Quan and Zekun Wang},
      year={2024},
      eprint={2412.15115},
      archivePrefix={arXiv},
      primaryClass={cs.CL},
      url={https://arxiv.org/abs/2412.15115},
}

@misc{bai2022traininghelpfulharmlessassistant,
      title={Training a Helpful and Harmless Assistant with Reinforcement Learning from Human Feedback},
      author={Yuntao Bai and Andy Jones and Kamal Ndousse and Amanda Askell and Anna Chen and Nova DasSarma and Dawn Drain and Stanislav Fort and Deep Ganguli and Tom Henighan and Nicholas Joseph and Saurav Kadavath and Jackson Kernion and Tom Conerly and Sheer El-Showk and Nelson Elhage and Zac Hatfield-Dodds and Danny Hernandez and Tristan Hume and Scott Johnston and Shauna Kravec and Liane Lovitt and Neel Nanda and Catherine Olsson and Dario Amodei and Tom Brown and Jack Clark and Sam McCandlish and Chris Olah and Ben Mann and Jared Kaplan},
      year={2022},
      eprint={2204.05862},
      archivePrefix={arXiv},
      primaryClass={cs.CL},
      url={https://arxiv.org/abs/2204.05862},
}

@misc{hendrycks2021measuringmathematicalproblemsolving,
      title={Measuring Mathematical Problem Solving With the MATH Dataset},
      author={Dan Hendrycks and Collin Burns and Saurav Kadavath and Akul Arora and Steven Basart and Eric Tang and Dawn Song and Jacob Steinhardt},
      year={2021},
      eprint={2103.03874},
      archivePrefix={arXiv},
      primaryClass={cs.LG},
      url={https://arxiv.org/abs/2103.03874},
}

@misc{lewkowycz2022solvingquantitativereasoningproblems,
      title={Solving Quantitative Reasoning Problems with Language Models},
      author={Aitor Lewkowycz and Anders Andreassen and David Dohan and Ethan Dyer and Henryk Michalewski and Vinay Ramasesh and Ambrose Slone and Cem Anil and Imanol Schlag and Theo Gutman-Solo and Yuhuai Wu and Behnam Neyshabur and Guy Gur-Ari and Vedant Misra},
      year={2022},
      eprint={2206.14858},
      archivePrefix={arXiv},
      primaryClass={cs.CL},
      url={https://arxiv.org/abs/2206.14858},
}

@misc{alpaca,
  author = {Rohan Taori and Ishaan Gulrajani and Tianyi Zhang and Yann Dubois and Xuechen Li and Carlos Guestrin and Percy Liang and Tatsunori B. Hashimoto },
  title = {Stanford Alpaca: An Instruction-following LLaMA model},
  year = {2023},
  publisher = {GitHub},
  journal = {GitHub repository},
  howpublished = {\url{https://github.com/tatsu-lab/stanford_alpaca}},
}

@misc{deepscaler2025,
  title={DeepScaleR: Surpassing O1-Preview with a 1.5B Model by Scaling RL},
  author={Luo, Michael and Tan, Sijun and Wong, Justin and Shi, Xiaoxiang and Tang, William Y and Roongta, Manan and Cai, Colin and Luo, Jeffrey and Li, Li Erran and Popa, Raluca Ada and others},
  year={2025},
  howpublished={\url{https://pretty-radio-b75.notion.site/DeepScaleR-Surpassing-O1-Preview-with-a-1-5B-Model-by-Scaling-RL-19681902c1468005bed8ca303013a4e2}},
  note={Notion Blog}
}

@inproceedings{ji-etal-2025-pku,
    title = "{PKU}-{S}afe{RLHF}: Towards Multi-Level Safety Alignment for {LLM}s with Human Preference",
    author = "Ji, Jiaming  and
      Hong, Donghai  and
      Zhang, Borong  and
      Chen, Boyuan  and
      Dai, Josef  and
      Zheng, Boren  and
      Qiu, Tianyi Alex  and
      Zhou, Jiayi  and
      Wang, Kaile  and
      Li, Boxun  and
      Han, Sirui  and
      Guo, Yike  and
      Yang, Yaodong",
    editor = "Che, Wanxiang  and
      Nabende, Joyce  and
      Shutova, Ekaterina  and
      Pilehvar, Mohammad Taher",
    booktitle = "Proceedings of the 63rd Annual Meeting of the Association for Computational Linguistics (Volume 1: Long Papers)",
    month = jul,
    year = "2025",
    address = "Vienna, Austria",
    publisher = "Association for Computational Linguistics",
    url = "https://aclanthology.org/2025.acl-long.1544/",
    doi = "10.18653/v1/2025.acl-long.1544",
    pages = "31983--32016",
    ISBN = "979-8-89176-251-0"
}

@inproceedings{he-etal-2024-olympiadbench,
    title = "{O}lympiad{B}ench: A Challenging Benchmark for Promoting {AGI} with Olympiad-Level Bilingual Multimodal Scientific Problems",
    author = "He, Chaoqun  and
      Luo, Renjie  and
      Bai, Yuzhuo  and
      Hu, Shengding  and
      Thai, Zhen  and
      Shen, Junhao  and
      Hu, Jinyi  and
      Han, Xu  and
      Huang, Yujie  and
      Zhang, Yuxiang  and
      Liu, Jie  and
      Qi, Lei  and
      Liu, Zhiyuan  and
      Sun, Maosong",
    editor = "Ku, Lun-Wei  and
      Martins, Andre  and
      Srikumar, Vivek",
    booktitle = "Proceedings of the 62nd Annual Meeting of the Association for Computational Linguistics (Volume 1: Long Papers)",
    month = aug,
    year = "2024",
    address = "Bangkok, Thailand",
    publisher = "Association for Computational Linguistics",
    url = "https://aclanthology.org/2024.acl-long.211/",
    doi = "10.18653/v1/2024.acl-long.211",
    pages = "3828--3850"
}

@misc{h4aime2024,
 author={{Hugging Face H4}},
 title={{AIME 2024 Dataset}},
 year={n.d.},
 url={https://huggingface.co/datasets/HuggingFaceH4/aime_2024},
 note={Dataset release; accessed September 15, 2026}
}

@misc{aimoamc2024,
 author={{AI-MO}},
 title={{AIMO Validation: AMC}},
 year={n.d.},
 url={https://huggingface.co/datasets/AI-MO/aimo-validation-amc},
 note={Dataset release; accessed September 15, 2026}
}

@misc{artessayrm2026,
 author={{Artessay}},
 title={{Qwen2.5-7B-SafeRLHF-RM}},
 year={n.d.},
 url={https://www.modelscope.cn/models/Artessay/Qwen2.5-7B-SafeRLHF-RM},
 note={Model card; accessed September 15, 2026}
}

@misc{artessaycm2026,
 author={{Artessay}},
 title={{Qwen2.5-7B-SafeRLHF-CM}},
 year={n.d.},
 url={https://www.modelscope.cn/models/Artessay/Qwen2.5-7B-SafeRLHF-CM},
 note={Model card; accessed September 15, 2026}
}
